\documentclass[10pt,twocolumn,letterpaper]{article}

\usepackage[pagenumbers]{cvpr}

\usepackage{graphicx}
\usepackage{multirow}
\usepackage{pifont}
\usepackage{algorithm}
\usepackage{algpseudocode}
\usepackage{placeins}  
\usepackage{amsthm}
\usepackage{tikz}
\usepackage{lipsum}

\usepackage{bm}
\usepackage{amssymb}
\usepackage{amsmath}
\usepackage[normalem]{ulem}

\def\p{{\mathrm{p}}}
\def\q{{\mathrm{q}}}

\DeclareMathOperator*{\argmax}{arg\,max}
\DeclareMathOperator*{\argmin}{arg\,min}

\definecolor{mygreen}{RGB}{38, 199, 149}
\newcommand{\mypar}[1]{\noindent\textbf{#1}\quad}

\definecolor{Gray}{gray}{0.9}

\usepackage{colortbl}
\newcommand{\gray}{\cellcolor{Gray}}

\newcommand*{\softmax}{\operatorname{Softmax}}
\newcommand*{\mult}{\operatorname{Mult}}

\def\Rbb{{\mathbb{R}}}

\def\Ebb{{\mathbb{E}}}

\def\cH{{\mathcal H}}

\def\cN{{\mathcal N}}

\def\bLambda{{\boldsymbol{\Lambda}}}

\def\bSigma{{\boldsymbol{\Sigma}}}

\def\bOmega{{\boldsymbol{\Omega}}}

\def\bp{{\mathbf p}}

\def\bt{{\mathbf t}}

\def\bw{{\mathbf w}}
\def\bx{{\mathbf x}}
\def\by{{\mathbf y}}

\def\bT{{\mathbf T}}

\def\bW{{\mathbf W}}
\def\bX{{\mathbf X}}
\def\bY{{\mathbf Y}}

\newcommand{\citeLP}{\cite{radford2021learning}}

\newtheorem*{claim_nn}{Claim}
\newtheorem{proposition}{Proposition}
\newtheorem*{proposition_nn}{Proposition}

\definecolor{cvprblue}{rgb}{0.21,0.49,0.74}
\usepackage[pagebackref,breaklinks,colorlinks,allcolors=cvprblue]{hyperref}

\definecolor{mygreen}{RGB}{38, 199, 149}

\def\paperID{11675} 
\def\confName{CVPR}
\def\confYear{2025}

\title{BayesAdapter: enhanced uncertainty estimation in CLIP few-shot adaptation}

\author{
Pablo Morales-Álvarez\thanks{Work done while visiting the International Laboratory on Learning Systems (ILLS) at ÉTS Montréal (Canada).}\\
Department of Statistics and Operations Research\\
University of Granada, Spain\\
\and
Stergios Christodoulidis\\
CentraleSupélec\\
Université Paris-Saclay, France\\
\and
Maria Vakalopoulou\\
CentraleSupélec\\
Université Paris-Saclay, France\\
\and
Pablo Piantanida\\
International Laboratory on Learning Systems (ILLS)\\
Quebec AI Institute (MILA)\\
CentraleSupélec, Université Paris-Saclay, France
\and
Jose Dolz\\
International Laboratory on Learning Systems (ILLS)\\
LIVIA, ÉTS Montréal, Canada
}

\begin{document}

\vspace{-2mm}

\maketitle

\vspace{-2mm}

\begin{abstract}  
    The emergence of large pre-trained vision-language models (VLMs) represents a paradigm shift in machine learning, with unprecedented results in a broad span of visual recognition tasks.
    CLIP, one of the most popular VLMs, has exhibited remarkable zero-shot and transfer learning capabilities in classification.
    To transfer CLIP to downstream tasks, adapters constitute a parameter-efficient approach that avoids backpropagation through the large model (unlike related prompt learning methods). 
    However, CLIP adapters have been developed to target discriminative performance, and the quality of their uncertainty estimates has been overlooked.
    In this work we show that the discriminative performance of state-of-the-art CLIP adapters does not always correlate with their uncertainty estimation capabilities, which are essential for a safe deployment in real-world scenarios.
    We also demonstrate that one of such adapters is obtained through MAP inference from a more general probabilistic framework. 
    Based on this observation we introduce BayesAdapter, which leverages Bayesian inference to estimate a full probability distribution instead of a single point, better capturing the variability inherent in the parameter space.
    In a comprehensive empirical evaluation we show that our approach obtains high quality uncertainty estimates in the predictions, standing out in calibration and selective classification.
    Our code will be publicly available upon acceptance of the paper. 
\end{abstract}

\section{Introduction} 

Recent advances in large pre-trained Vision-Language Models (VLMs), such as CLIP \cite{radford2021learning},
have fused visual perception with language comprehension, reshaping the computer vision field through impressive zero-shot and generalization capabilities. These models effectively tackle tasks without the need for specialized task-specific training, enabling a wide array of visual recognition problems, from content-based image retrieval \cite{sain2023clip} to segmentation \cite{hajimiri2024pay,wang2025sclip}.
In this work we focus on the classification problem \cite{radford2021learning}. 

While the zero-shot capability of CLIP is highly valuable, its performance at a particular task can be enhanced if training data are available. 
\textit{Prompt Learning} approaches
introduce 
a learnable textual prompt that
can be optimized to fit the training data \cite{zhou2022conditional,zhou2022coop}.
However, this requires backpropagation through CLIP textual encoder, which is computationally demanding and precludes \textit{black-box} adaptation.
Alternatively, \emph{adapters} \cite{tipA_zhang2021tip, silva2024closer, lppp_lp24} have emerged as a parameter efficient strategy, as they introduce some learnable parameters to only transform the extracted features, keeping both visual and textual encoders frozen.

CLIP adapters have been designed to target discriminative performance.
Indeed, most of the literature focuses on the test accuracy achieved after adaptation \cite{radford2021learning, tipA_zhang2021tip, CrossModal_lin2023crossmodal, TR_yu2023task, lppp_lp24, silva2024closer}.
However, in safety-critical scenarios, where CLIP is becoming increasingly popular \cite{liang2022effective,liu2023clip,shakeri2024few}, it is essential to consider the uncertainty (or confidence) on the predictions.
Firstly, confidence scores should be \emph{well calibrated} \cite{chuan_calibration}. 
This means that the confidence score should reflect the true accuracy: if we have several test samples with confidence of $90\%$, around $90\%$ of them should be correct.
Calibration thus avoids both overconfident predictions, which would lead to a dangerous system, and underconfident ones, which would yield a too conservative system \cite{Nixon_2019_CVPR_Workshops}.
Secondly, confidence scores should allow for making reliable predictions at a desired confidence level (\emph{selective classification}).
Although different metrics have been proposed to measure prediction confidence \cite{liu_2020, dadalto_relu}, we stick to maximum softmax probability for being the most popular one. 

In this work we evaluate current CLIP adapters in terms of uncertainty estimation, revealing that their uncertainty estimation capabilities does not always correlate with their discriminative performance.
Indeed, some of the best methods in test accuracy consistently fail in terms of other metrics involving the confidence score, such as calibration and coverage at high confidence. 
Then, we show that the methodology behind a state-of-the-art method in discriminative power, CLAP \cite{silva2024closer}, can be cast as maximum-a-posteriori (MAP) inference of a general probabilistic adaptation framework. 
This implies that CLAP relies solely on a \emph{point estimate} of the parameters, resulting in suboptimal handling of uncertainty.

To alleviate this issue, we propose to leverage Bayesian inference instead of MAP \cite{BDL_2024}.
By doing so, one can estimate a probability distribution over the parameters that are consistent with the observed data. 
Nevertheless, Bayesian inference cannot be performed in closed-form in our probabilistic model. Hence, we resort to the variational Bayes approach to approximate the true posterior distribution \cite{blei2017variational, morales2020activation}. 
By estimating a probability distribution instead of a point estimate, we can better handle uncertainty quantification in the parameters. Interestingly, we will show that this leads to enhanced uncertainty estimation in the predictions.

In a comprehensive empirical evaluation involving eleven different datasets, we show that our BayesAdapter outperforms its deterministic counterpart in different problems related to uncertainty estimation, such as calibration ($\sim$2.5\% gain in ECE) and selective classification at high confidence ($\sim$6-9\% gain in test set coverage at 99\% confidence); while being competitive in discriminative performance ($\sim$0.5-0.7\% below in accuracy).
These advantages persist when compared to six more state-of-the-art adapters.

\noindent Our key contributions can be summarized as:
\begin{itemize}
    \item Analyzing current CLIP adapters beyond accuracy, in terms of the quality of their uncertainty estimates. This reveals interesting insights about current adapters,
    showing that their discriminative performance does not always correlate with their uncertainty estimation capabilities (in both calibration and selective classification). 
    \item We demonstrate that the methodology behind the best performing method in terms of accuracy, CLAP \cite{silva2024closer}, is a maximum-a-posteriori (MAP) estimation of a general probabilistic adaptation framework with a Gaussian prior. In particular, MAP inference implies that only a point estimate of the parameters is leveraged.
    \item By using Bayesian inference (instead of MAP), we introduce a novel approach that leverages a probability distribution over the parameters, instead of a single point. This enables a richer uncertainty quantification in the parameter space.
    \item We show empirically that this better uncertainty quantification at parameter-level translates into the predictions, both in terms of calibration and selective classification, while discriminative power remains competitive. We provide a comprehensive evaluation with seven recent CLIP adapters baselines and eleven datasets with diverse complexity, task granularity and number of categories. 
\end{itemize}

\section{Related work}

\mypar{Few-shot adaptation of VLMs.} While VLMs are well-known for their promising zero-shot performance, few-shot adaptation of these models, and more particularly of CLIP, has garnered considerable interest to accommodate these models to novel tasks. The prevailing literature on this task is dominated by \textit{Prompt Learning} \cite{zhou2022conditional,zhou2022coop,khattak2023maple,zhu2023prompt} and \textit{Adapter-based} \cite{gao2021clip,tipA_zhang2021tip,silva2024closer,lppp_lp24,CrossModal_lin2023crossmodal,TR_yu2023task}  methods.
Prompt Learning integrates a set of continuous learnable tokens into the text original prompt, which are provided as the input for the CLIP text encoder. Then, the whole CLIP model remains frozen, whereas the learnable tokens are optimized based on the few-shot samples provided. Nevertheless, these techniques backpropagate gradients over the entire model, requiring long training processes and precluding \textit{black-box} adaptation. In contrast, adapter-based approaches have emerged as a significantly more efficient alternative, as only a small set of parameters
must be optimized. For example, state-of-the-art approaches \cite{silva2024closer,lppp_lp24} fine-tune just a linear layer on top of the CLIP features.

\mypar{Uncertainty estimation in VLMs.} Despite the rapid adoption of these models in critical areas, the understanding of their uncertainty estimation capabilities remains a relatively unexplored area.
Albeit a few attempts have been made in this direction \cite{yoonc,tuempirical,oh2023towards,wangopen,murugesan2024robust, khan2024consistency,upadhyay2023probvlm}, they present important differences with our work. For example, C-TPT \cite{yoonc} and DAC \cite{wangopen} study the miscalibration problem on prompt learning strategies, whereas \cite{oh2023towards} is presented in the full fine-tuning scenario. 
These settings require backpropagation through the textual encoder, which is expensive and precludes black-box adaptation.
Furthermore, \cite{tuempirical} assess the impact of Temperature Scaling on zero-shot VLM predictions, but do not investigate its effect after few-shot adaptation.
A straightforward logit normalization strategy was presented in \cite{murugesan2024robust} to calibrate adapters in the context of domain shift. However, only one dataset (ImageNet) is used for the adapters, and the approach does not perform well in our more extensive evaluation (see Sec. \ref{sec:app_other_tab_figs} in the Appendix).
Selective classification based on uncertainty was recently studied in the context of VLMs for visual question answering \cite{khan2024consistency}, but no adaptation to new tasks was performed.
In \cite{upadhyay2023probvlm}, the authors introduce probability distributions for the embeddings of the VLMs, but they do it in the context of refining the vision-language contrastive learning, and not for transfer to downstream tasks.

\mypar{Bayesian inference in deep learning and large VLMs.}
Bayesian methods have been extensively leveraged in combination with deep neural networks, leading to so-called Bayesian deep learning \cite{izmailov2021bayesian, kapoor2022uncertainty, BDL_2024}.
Regarding VLMs, several investigations have employed Bayesian methods in different directions.
CLIPScope proposes a Bayesian scoring in the context of zero-shot classification \cite{fu2024clipscope}.
In \cite{wang2024bayesian}, the authors leverage a Bayesian solution to select the most appropriate samples to fine-tune large AI models, and \cite{miao2024bayesian} introduces a Bayesian framework based on Gaussian Processes to combine knowledge form different large models such as CLIP \cite{radford2021learning}, DINO \cite{caron2021emerging} and MoCo \cite{chen2021empirical}.
A Bayesian cross-modal image-text alignment method has been recently proposed in \cite{zhu2023bayesian} to improve the discriminative generalization ability under distribution shift.
Bayesian prompt learning is proposed in \cite{Derakhshani_2023_ICCV}, but it requires backpropagation through the VLM.
Despite these related efforts, to the best of our knowledge, Bayesian inference has not been leveraged in the context of CLIP adapters before.

\section{Methodology}\label{sec:theory}

\subsection{Preliminaries}
\mypar{Zero-shot classification with CLIP.}
Given a classification problem, CLIP allows zero-shot (ZS) classification, that is, having no access to training data \cite{radford2021learning}.
Indeed, CLIP is composed of one visual ($\psi_v$) and one textual ($\psi_t$) encoders, which have been trained on a large corpus of image-text pairs via contrastive learning, enforcing visual and text embeddings to be close for related image-text pairs, and far away for not related ones (distance is measured in the $\ell_2$-normalized embedding space through the cosine similarity) \cite{fan2024improving}. 
Therefore, for a new test image $\bx_*$ and $C$ possible classes, one can measure the similarity between the visual embedding $\psi_v(\bx_*)\in\Rbb^{1\times D}$ and each one of the class text embeddings $\psi_t(T_c)\in\Rbb^{1\times D}$ by computing $\psi_v(\bx_*)\cdot\psi_t(T_c)^\top$, and then predicting the class that corresponds to the highest similarity. 
Here, each $T_c$ is a textual description of the class (usually as simple as \textsc{\textcolor{gray}{``An image of a [cls$_c$]''}}, where \textsc{cls$_c$} is the name of the $c$-th class). The vectors $\psi_t(T_c)$ are called \emph{class prototypes}.

\mypar{CLIP adaptation in the presence of few-shots.} Even though ZS classification provides good results without requiring supervision, its performance can be significantly enhanced in the presence of training data $(\bX,\bY)=\{(\bx_n,\by_n)\}_{n=1}^{N}$\cite{silva2024closer,kgcoop23}. 
Although this can be done with a training set of any size, pre-trained VLMs are particularly interesting when training data is scarce.
Therefore, we follow the standard few-shot formulation, where the training set consists of $K$ samples (or shots) for each class, and thus $N=K\times C$.
Finally, in line with the assumption of training data scarcity and following previous work, we assume there is no access to a validation set to fine-tune model hyperparameters or to perform early-stopping \cite{silva2024closer, murugesan2024robust}.

\mypar{CLAP: current state-of-the-art adapter in discriminative performance.} Considering a realistic validation-free scenario, \cite{silva2024closer} shows that existing CLIP adapters significantly degrade their performance when they have no access to task-specific validation sets for model selection.
Then, they propose a simple adapter based on linear probing, referred to as Class Adaptive Constraint for Linear Probing (CLAP), which achieves top discriminative performance.

CLAP is a linear probing approach where the trainable linear layer is initialized with the class zero-shot embeddings (instead of randomly). Moreover, it includes a regularization term to penalize when it deviates much from its initialization.
Formally, CLAP optimizes this objective:
\begin{equation}\label{eq:CLAP_obj}
    \min_{\bW}\left[\sum_{n=1}^N \cH(\by_n, \hat\by_n)+\sum_{c=1}^C \lambda_c ||\bw_c - \bt_c||^2\right],
\end{equation}
where
$\cH$ is the cross-entropy (CE) loss,
$\lambda_c$ are class-dependent weights to control the regularization term, $\bt_c=\psi_t(T_c)$ are 
the class prototypes,
$\bW=(\bw_1^\top,\dots,\bw_C^\top)$ is the learnable $D\times C$ matrix initialized to $\bT=(\bt_1^\top,\dots,\bt_C^\top)$,
and the predicted probabilities $\hat\by_n$ are obtained from the input $\bx_n$ and $\bW$ through a simple softmax linear model:
\begin{equation}
\hat\by_n = \softmax(\psi_v(\bx_n)\cdot \bW).
\end{equation}

\subsection{Our novel BayesAdapter}

\mypar{CLAP is obtained as MAP inference for a general probabilistic adapter.}
From a general probabilistic perspective, a training loss like the one in~\eqref{eq:CLAP_obj}, consisting of a CE term and a $\ell_2$ regularizer, can be viewed as performing maximum-a-posteriori (MAP) inference in a probabilistic model with: i) a Gaussian prior over its parameters and ii) a likelihood function based on the multinomial distribution.

In this scenario, the probabilistic model is defined as:
\begin{align}\label{eq:prob_model_prior}
    \p(\bW) &= \cN(\bW| \bT, \bLambda), \\
    \p(\by|\bW, \bx) &= \mult(\by|\softmax(\psi_v(\bx)\cdot \bW)),\label{eq:prob_model_lik}
\end{align}
where $\bLambda=2\cdot\mathrm{diag}(\lambda_1^{-1},\dots,\lambda_1^{-1},\dots,\lambda_C^{-1},\dots,\lambda_C^{-1})$ is a diagonal matrix with each $\lambda_c$ repeated $D$ times (the dimensionality of the embedding space), and recall that $\bT=(\bt_1^\top,\dots,\bt_C^\top)$.
We can then state the sought result, whose derivation is in the Appendix, Sec. \ref{sec:app_theoretical_der}. 

\begin{proposition}\label{prop:1}
Given training data $(\bX,\bY)$, maximizing the (log) posterior probablity $\p(\bW|\bX,\bY)$ for the model in eqs.\eqref{eq:prob_model_prior}-\eqref{eq:prob_model_lik} is equivalent to minimizing the loss in eq.~\eqref{eq:CLAP_obj}.
\end{proposition}

\begin{figure}
    \centering
    \includegraphics[width=1.05\columnwidth]{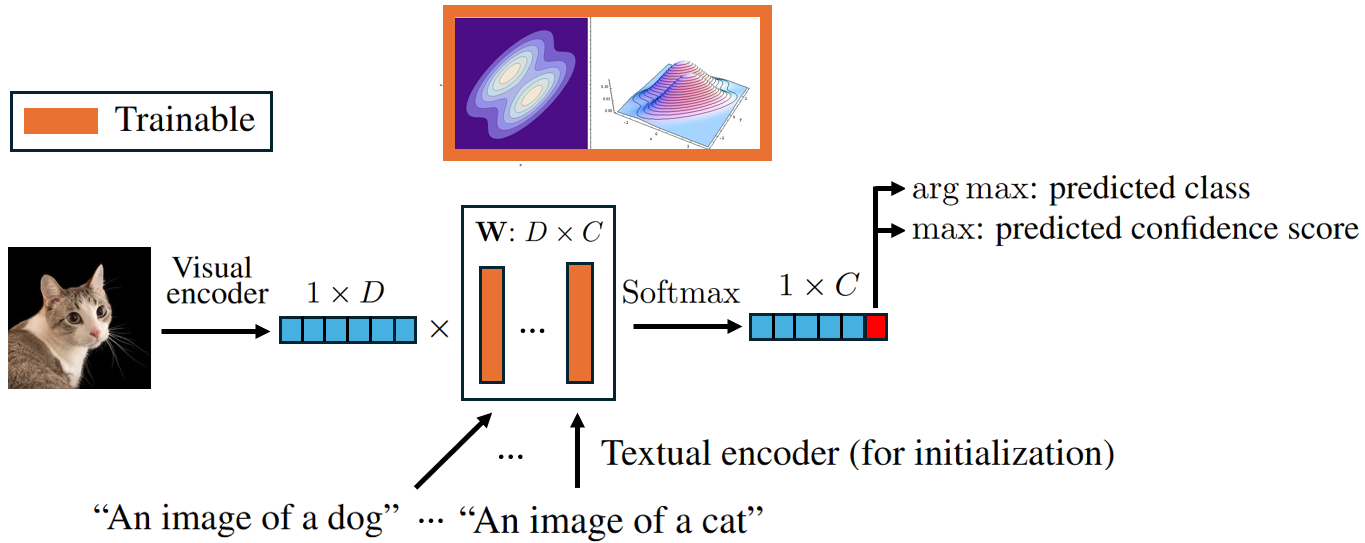}
    \caption{Graphical representation of the novel BayesAdapter. Whereas CLAP estimates $\bW$ through a single point estimate, BayesAdapter estimates a probability distribution over it.}
    \label{fig:enter-label}
\end{figure}

\mypar{BayesAdapter: leveraging Bayesian inference.}
By performing MAP inference, CLAP estimates a single $\bW$. However, this does not capture the uncertainty encoded in the posterior distribution $\p(\bW|\bX,\bY)$. 
Therefore, we propose to use Bayesian inference to leverage a probability distribution instead of a single value \cite{BDL_2024, BDL_2020}. 
In the experiments we will see that this better quantification of uncertainty at the parameter level translates into better uncertainty estimates in the predictions too. 

Notice that the posterior distribution cannot be obtained in closed-form, since the Gaussian and Multinomial distributions in eqs.\eqref{eq:prob_model_prior}-\eqref{eq:prob_model_lik} are not conjugate. Therefore, we resort to approximate Bayesian inference.
Specifically, we consider the Variational Bayes approach, which is amenable to mini-batch training and has been applied in many supervised and unsupervised probabilistic models \cite{blei2017variational, morales_tpami2022, kingma2013auto, kingma2021variational}.

In variational inference, one considers an approximate posterior distribution $\q_\alpha(\bW)$ that depends on some parameters $\alpha$, and minimizes the Kullback-Leibler (KL) divergence between $\q_\alpha(\bW)$ and the true posterior $\p(\bW|\bX,\bY)$ \cite{zhang2018advances}.
The KL divergence is commonly used as a distance between distributions: it is non-negative, and equals zero only if both distributions are identical.
Minimizing the KL divergence is equivalent to minimizing the negative Evidence Lower Bound (ELBO) \cite{blei2017variational, CASTROMACIAS2024104115}, which is expressed as:
{\small
\begin{equation}\label{eq:ours_obj}
    \min_{\alpha}\left[\sum_{n=1}^N\left( \Ebb_{\q_\alpha(\bW)}\cH(\by_n, \hat\by_n)\right)\!+\!\mathrm{KL}(\q_\alpha(\bW)||\p(\bW))\right].
\end{equation}
}
This equivalence is derived step-by-step in the Appendix, Sec.~\ref{sec:app_theoretical_der}.
It is interesting to observe the similarity between eq.~\eqref{eq:ours_obj} and CLAP training objective, i.e., eq.~\eqref{eq:CLAP_obj}.
In both cases, there is a data-fidelity term given by the CE loss and a regularizer that penalizes parameters moving far away from the zero-shot prototypes. 
Yet, whereas a single point estimate $\bW$ is sought in eq.~\eqref{eq:CLAP_obj}, a probability distribution over $\bW$ is sought in eq.~\eqref{eq:ours_obj} (by optimizing the parameters $\alpha$). 

In our case, we consider a Gaussian posterior distribution with a block-diagonal covariance matrix, similar to the prior. Namely, we consider $\q(\bW)=\cN(\bW|\bOmega, \bSigma)$, where $\bOmega\in\Rbb^{D\times C}$ and $\bSigma=\mathrm{diag}(\sigma_1^2,\overset{D}{\cdots},\sigma_1^2,\cdots,\sigma_C^2,\overset{D}{\cdots},\sigma_C^2)$ are the parameters $\alpha$ to be learned.
Therefore, in addition the the posterior mean $\bOmega$, we need to estimate class-wise variances $\sigma_1^2,\dots,\sigma_C^2$.
This choice of posterior allows us to evaluate $\eqref{eq:ours_obj}$ following standard approaches in variational inference \cite{zhang2018advances}: the expectation can be computed through Monte Carlo (MC) sampling by leveraging the reparametrization trick of the Gaussian distribution \cite{kingma2013auto}, and the KL divergence between two Gaussians has a closed-form expression (efficiently implemented in Pytorch). 

The training process is summarized in Algorithm~\ref{alg:BayesAdapter}.
Note that backpropagation in step 6, which is w.r.t. the Gaussian distribution parameters, can be performed straightforwardly in automatic differentiation frameworks such as Pytorch because we are using the reparametrization trick, originally introduced for VAEs \cite{kingma2013auto}.

\begin{algorithm}
\caption{Training BayesAdapter.\label{alg:BayesAdapter}}
\begin{algorithmic}[1]
\small
\Require 
(1) Training data $(\bX, \bY)$;
(2) Prior distribution parameters $(\bT,\bLambda)$;
(3) Posterior distribution initializations $(\bOmega(0), \bSigma(0))$; 
(4) Other hyperparameters: number of epochs $N_e$, batch size \( b_s \), number of samples for MC sampling \(s_{\mathrm{MC}}\).

\State Compute number of batches \( B \) in terms of \( N_e \) and \( b_s \)
\State \( t \gets 0 \)
\For{epoch = 1, \dots, \( N_e \)}
    \For{batch = 1, \dots, \( B \)}
        \State Calculate the neg-ELBO in eq.~\eqref{eq:ours_obj}, using current parameters $\bOmega(t), \bSigma(t)$ and current batch. The MC sampling uses \( s_{\mathrm{MC}} \) samples.
        \State Obtain gradients $\nabla_\bOmega$ and $\nabla_\bSigma$ of neg-ELBO at the current values $\bOmega(t), \bSigma(t)$.
        \State Compute $\bOmega(t+1), \bSigma(t+1)$ based on the optimizer, gradients, and current values.
        \State \( t \gets t + 1 \)
    \EndFor
\EndFor

\State \textbf{Output:} Learned parameters $\hat\bOmega, \hat\bSigma$.
\end{algorithmic}
\end{algorithm}

\noindent \textit{Inference.} To predict on a new image $\bx_*$, we use the learned posterior $\q(\bW)$ and eq.~\eqref{eq:prob_model_lik} to calculate $\Ebb_{\q(\bW)} \mult(\by|\softmax(\psi_v(\bx_*)\cdot \bW))$.
As this cannot be obtained in closed-form, we perform MC integration to yield a final set of softmax test probabilities $\bp = (p_1,\dots,p_C)$, $\sum_c p_c = 1$. The predicted class is obtained as $\argmax(\bp)$, and the confidence score (resp., uncertainty score) is $\max(\bp)$ (resp., $1-\max(\bp)$).

\section{Experiments} 


\begin{table*}[ht]
\footnotesize
    \centering
\caption{\label{tab:acc_cal_all}\textbf{Comparison in terms of calibration and discriminative performance} across two popular visual backbones.
   Results are averaged over all 
   configurations (11 datasets, 6 options for number of shots, and 3 seeds). The evolution w.r.t. to the number of shots is analyzed in Fig.~\ref{fig:evol_shots}, and detailed 
   numerical values are in the Appendix, Sec.~\ref{sec:app_full_results}. For each metric: best result in bold; second best underlined.}
    \begin{tabular}{lcccccc}
\toprule
& \multicolumn{3}{c}{ResNet50} & \multicolumn{3}{c}{ViT-16}\\
\cmidrule(lr){2-4}
\cmidrule(lr){5-7}
& Accuracy ($\uparrow$) & ECE ($\downarrow$) & AECE ($\downarrow$)  & Accuracy ($\uparrow$) & ECE ($\downarrow$) & AECE ($\downarrow$) \\
\midrule
LP$_\text{ICML'21}$ & 54.086 & 24.307 & 24.297 & 62.755 & 20.222 & 20.196 \\
TipA$_\text{ECCV'22}$ & 62.824 & 7.772 & 7.740 & 64.435 & 31.321 & 31.306 \\
TipA-f-$_\text{ECCV'22}$ & 67.661 & 5.865 & 5.799 & 66.260 & 29.379 & 29.365 \\
CrossModal$_\text{CVPR'23}$ & 68.987 & \underline{4.611} & \underline{4.601} & 75.881 & \underline{3.885} & \underline{3.831} \\
TaskRes$_\text{CVPR'23}$ & 69.811 & 5.729 & 5.735 & \underline{76.851} & 4.475 & 4.460 \\
LP++$_\text{CVPR'24}$ & \underline{69.846} & 13.812 & 13.821 & 75.998 & 8.353 & 8.353 \\
CLAP$_\text{CVPR'24}$ & \textbf{70.182} & 6.592 & 6.572 & \textbf{76.980} & 5.351 & 5.323 \\
 \gray BayesAdapter (\textit{Ours})
& \gray 69.476 & \gray \textbf{4.151} & \gray \textbf{4.098} & \gray 76.420 & \gray \textbf{3.835} & \gray \textbf{3.796} \\
\bottomrule
\end{tabular}
\end{table*}

\begin{figure*}[ht]
    \centering
    \begin{subfigure}{0.22\textwidth}
        \includegraphics[width=\linewidth]{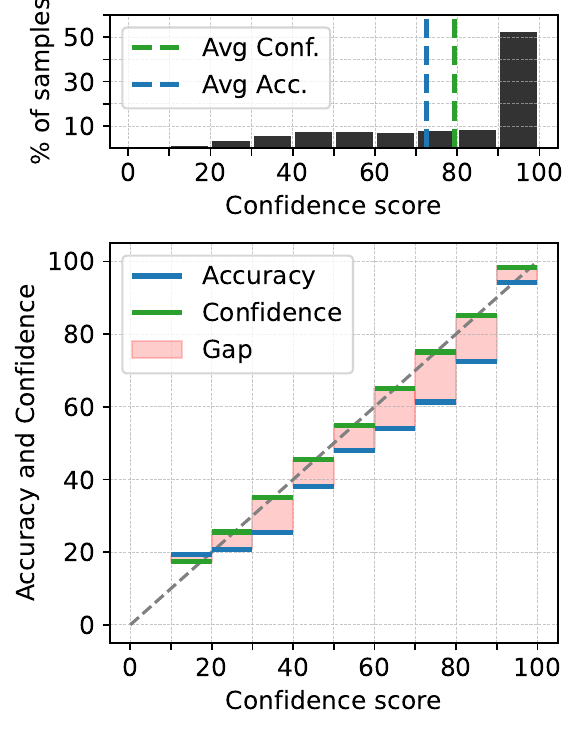}
        \caption{TipA-f-}
    \end{subfigure}
    \hfill
    \begin{subfigure}{0.22\textwidth}
        \includegraphics[width=\linewidth]{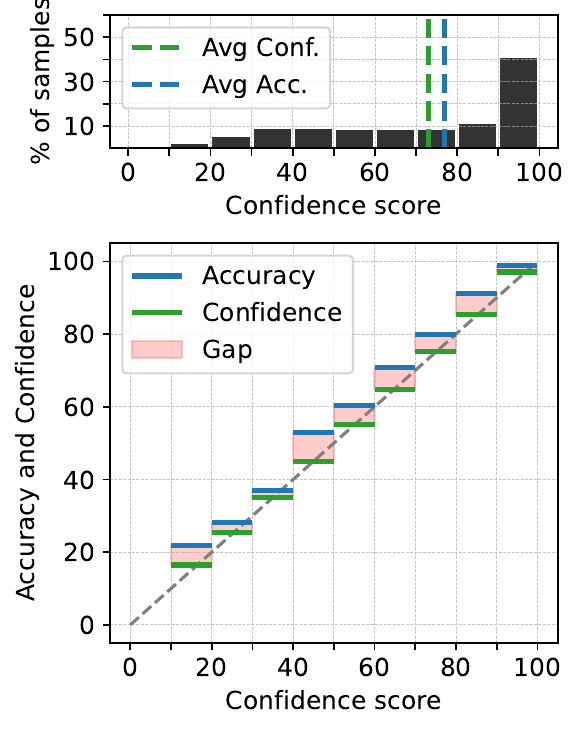}
        \caption{CrossModal}
    \end{subfigure}
    \hfill
    \begin{subfigure}{0.22\textwidth}
        \includegraphics[width=\linewidth]{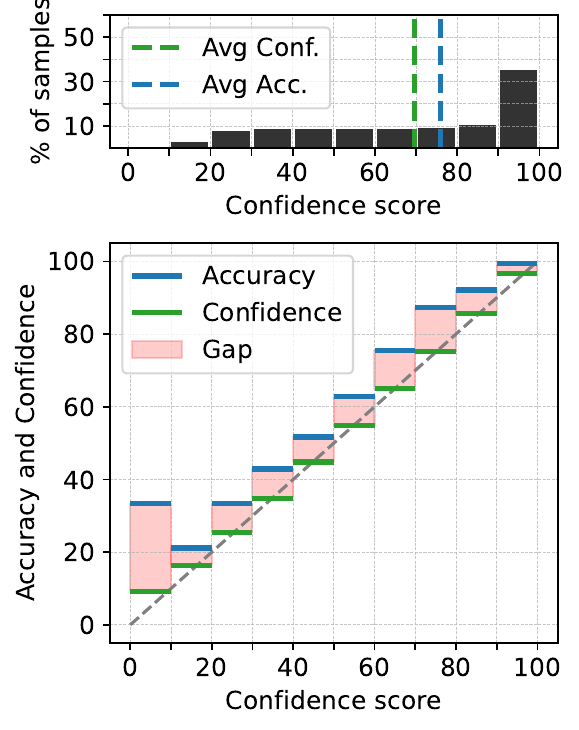}
        \caption{TaskRes}
    \end{subfigure}
    \hfill
    \begin{subfigure}{0.22\textwidth}
        \includegraphics[width=\linewidth]{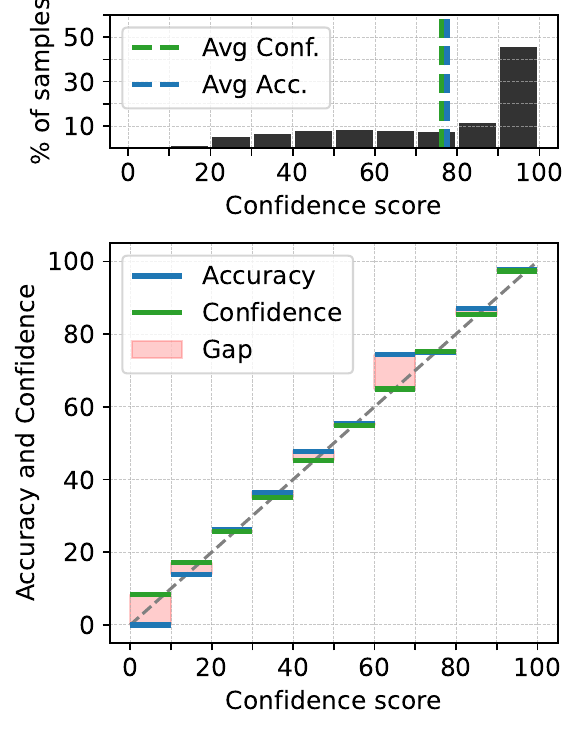}
        \caption{BayesAdapter (ours)}
    \end{subfigure}
    \caption{\textbf{Calibration plots} for the four best methods in terms of ECE in Table \ref{tab:acc_cal_all} (ResNet50 backbone). A full figure with all the methods is shown in the Appendix, Sec.~\ref{sec:app_other_tab_figs}. 
    In each case, the lower subplot depicts the accuracy and average confidence for samples in each one of the ten bins (from 0\% to 100\% of confidence score by steps of 10\%). Ideally, the gap between them should be zero. The upper plot shows the proportion of samples in each bin, along with the average confidence and accuracy in the whole test set. 
    \label{fig:calib_plots}
    }
\end{figure*}

\subsection{Experimental setting}\label{sec:exp_setting}

\noindent \textbf{Datasets.}
We follow prior literature on CLIP apdaters \cite{TR_yu2023task,gao2021clip,tipA_zhang2021tip,silva2024closer} and utilize 11 diverse datasets, including Imagenet \cite{deng2009imagenet}, Caltech101 \cite{caltech}, OxfordPets \cite{oxfordpets}, StanfordCars \cite{stanfordcars}, Flowers102 \cite{flowers102}, Food101 \cite{food101}, FGVCAircraft \cite{aircraft}, SUN397 \cite{sun397}, DTD \cite{dtd}, EuroSAT \cite{eurosat}, and UCF101 \cite{ucf101}.
These datasets cover a wide range of computer vision classification tasks, from general objects to fine-grained categories in specialized applications. To train all the adapters, $K$ shots ($K\in \{1, 2, 4, 8, 16, 32\}$) are randomly selected for each class.
For evaluation, we use the test sets provided in each dataset, following the same splits as in \cite{zhou2022coop,TR_yu2023task,silva2024closer}.

\noindent \textbf{Baselines and adaptation protocol.} 
We compare against seven recent and popular CLIP adapters:
standard Linear Probing (LP) \citeLP,
TIP-Adapter (TipA) and TIP-Adapter-full (TipA-f-) \cite{tipA_zhang2021tip},
TaskRes \cite{TR_yu2023task},
CrossModal \cite{CrossModal_lin2023crossmodal},
LP++ \cite{lppp_lp24}, 
and CLAP \cite{silva2024closer}.
To meet with real-world requirements, we follow the strict few-shot adaptation protocol presented in \cite{silva2024closer}, where no additional validation or test samples are accessible to find the best case-specific configuration for each method, and therefore the hyperparameters remain fixed across all tasks.

\noindent \textbf{General implementation details.}
CLIP pre-trained features are obtained with two visual backbones: ResNet-50 \cite{resnet} and ViT-B/16 \cite{dosovitskiy2020vit}. Unless otherwise specified, we resort to ResNet-50 in the ablation studies.
Data augmentation (random zoom, crops, and flips) is applied during feature extraction following \cite{TR_yu2023task,silva2024closer}, with 20 augmentations per support sample.
Furthermore, we used the same text prompts per dataset as in \cite{TR_yu2023task,silva2024closer}.
All the adapters are trained under the same setting: 300 epochs, batch size of 256, and SGD optimizer with momemtum 0.9 and learning rate of 0.1 (which decreases during training following a cosine decay scheduler). We run all the experiments on the same GPU (NVIDIA GeForce RTX 3090) with three seeds, and average results are reported (full numerical values with standard errors can be found in the Appendix, Sec.~\ref{sec:app_full_results}).
Our code will be publicly available upon acceptance of the paper. 

\mypar{Implementation details specific to BayesAdapter.}
Regarding the initializations and hyperparameters that are specific to BayesAdapter, the prior mean $\bT$ is set to the ZS prototypes, since we are inspired by the same CLAP rationale of leveraging the prior knowledge in the textual encoder \cite{silva2024closer}.
The initialization of the posterior mean $\bOmega$ is $\bOmega(0)=\bT$.
Regarding the prior variance $\bLambda$, the standard deviation for all classes is set to $0.01$. We choose this value based on the average empirical performance of the ZS classification, and fix it throughout the experimentation.
In the ablation study, see Table \ref{tab:ablation}, we analyze other values. The posterior variance is initialized to the prior variance value.
The number of MC samples is set to $s_{\mathrm{MC}}=3$ in all the experiments. Other values are analyzed in Table \ref{tab:ablation}.
Following common practice in variational inference \cite{huang2018improving}, we utilize linear KL annealing during training.

\subsection{Experimental results}


\begin{table*}
\footnotesize
    \centering
\caption{\label{tab:coverage_sample_level}
    \textbf{Comparison in terms of test set coverage at different levels of confidence}.
    This measures the \% of test samples that have confidence above the threshold, provided the method is reliable at such confidence level (i.e. the accuracy in the selected subset is above the requested confidence). A value of {\color{red}\ding{55}} means that the method is not reliable.
    Best result in bold; second best underlined.}
    \begin{tabular}{lrrrrrrrrrr}
\toprule
 & \multicolumn{5}{c}{ResNet50} & \multicolumn{5}{c}{ViT-16} \\
 \cmidrule(lr){2-6}
 \cmidrule(lr){7-11}
 & \multicolumn{5}{c}{($\uparrow$) Test set coverage at confidence...} & \multicolumn{5}{c}{($\uparrow$) Test set coverage at confidence...} \\
 & 99\% & 95\% & 90\% & 85\% & 80\% & 99\% & 95\% & 90\% & 85\% & 80\% \\
\midrule
LP$_\text{ICML'21}$ & {\color{red} \ding{55}} & {\color{red} \ding{55}} & {\color{red} \ding{55}} & {\color{red} \ding{55}} & {\color{red} \ding{55}} & {\color{red} \ding{55}} & {\color{red} \ding{55}} & {\color{red} \ding{55}} & {\color{red} \ding{55}} & {\color{red} \ding{55}} \\
TipA$_\text{ECCV'22}$ & {\color{red} \ding{55}} & {\color{red} \ding{55}} & 28.956 & 32.857 & 36.172 & {\color{red} \ding{55}} & {\color{red} \ding{55}} & {\color{red} \ding{55}} & {\color{red} \ding{55}} & {\color{red} \ding{55}} \\
TipA-f-$_\text{ECCV'22}$ & {\color{red} \ding{55}} & \underline{24.480} & 30.307 & 34.457 & 37.890 & {\color{red} \ding{55}} & {\color{red} \ding{55}} & {\color{red} \ding{55}} & {\color{red} \ding{55}} & {\color{red} \ding{55}} \\
CrossModal$_\text{CVPR'23}$ & \underline{10.065} & 22.973 & \underline{31.314} & \underline{37.236} & \underline{42.238} & \underline{17.272} & \underline{34.318} & \underline{43.518} & \underline{49.597} & \underline{54.436} \\
TaskRes$_\text{CVPR'23}$ & 8.346 & 20.656 & 28.907 & 34.964 & 40.049 & 14.894 & 31.900 & 41.501 & 47.923 & 53.003 \\
LP++$_\text{CVPR'24}$ & 2.222 & 7.453 & 12.415 & 16.598 & 20.408 & 6.046 & 17.902 & 25.798 & 31.376 & 35.886 \\
CLAP$_\text{CVPR'24}$ & 7.103 & 19.035 & 27.529 & 33.834 & 39.080 & 12.534 & 29.367 & 39.623 & 46.510 & 51.930 \\
 \gray BayesAdapter$_\text{Ours}$ & \gray \textbf{13.423} & \gray\textbf{27.992} & \gray\textbf{36.749} & \gray\textbf{42.861} & \gray\textbf{47.868} & \gray\textbf{21.906} & \gray\textbf{39.929} & \gray\textbf{49.139} & \gray\textbf{55.122} & \gray\textbf{59.846} \\
\bottomrule
\end{tabular}
\end{table*}

\subsubsection{Calibration}

\mypar{Concept and metrics.}
A model is well-calibrated if the confidence score of its predictions is a good proxy of its accuracy.
This is an important aspect for safe deployment in real-world scenarios, since a poorly calibrated model could lead to overconfident predictions or be too conservative to make confident predictions. 
The most popular metric is the Expected Calibration Error (ECE) \cite{chuan_calibration}, which divides the test samples in $B$ bins of the same length in terms of confidence score and computes
$\mathrm{ECE}=\sum_{b=1}^B q_b |\mathrm{acc}_b-\mathrm{conf}_b|$, where $\mathrm{acc}_b$ and $\mathrm{conf}_b$ are the average accuracy and confidence in the $b$-th bin, and $q_b$ is the proportion of samples in that bin.
The adaptive ECE (AECE) is a popular variant that adapts bins size according to the number of predictions in each range, ensuring that each bin contains approximately the same number of samples. This addresses potential biases when the standard ECE may be skewed by having bins with few or no samples \cite{Nixon_2019_CVPR_Workshops}. We use $B=10$ bins.

\mypar{Evaluation of the compared methods.}
Table \ref{tab:acc_cal_all} reports ECE and AECE. As theoretically expected, BayesAdapter outperforms its deterministic counterpart CLAP in these uncertainty-related metrics, and achieves the best results among all methods across both visual backbones.
To further illustrate the calibration behavior, Figure \ref{fig:calib_plots} depicts calibration plots for the four best methods in terms of ECE, using ResNet-50. The full figure is in the Appendix, Sec.~\ref{sec:app_other_tab_figs}.
We observe a shared pattern across all datasets: whereas earlier methods LP, TipA and TipA-f- are overconfident (their confidence is above their accuracy), more recent approaches CrossModal, TaskRes, LP++ and CLAP are underconfident.
The proposed BayesAdapter obtains the best calibration through its principled quantification of uncertainty.

\subsubsection{Selective classification at high confidence}

\mypar{Concept and metrics.}
When deploying a system in safety-critical scenarios, it is important to have a reliable confidence score to select which cases can be automatically classified with a high confidence \cite{dadalto_relu}. And, provided the score is reliable, one seeks maximum coverage of the test set. 
For example, if we want to predict at 99\% confidence, we need that at least 99\% of the selected samples are correctly predicted.
Nonetheless, given that, we would prefer a method that selects 35\% of the test set over another one that only selects 1\% of it, which would be ``reliable but too conservative'' compared to the former. 
Following previous work \cite{SC_CCL-SC, sc_NIPS2017_4a8423d5, sc_xia2024understanding}, in the sequel we will refer to a method as \emph{reliable at confidence level X\%} if the accuracy on the selected subset is greater or equal than X\%.
We will evaluate reliability and coverage at different confidence levels.

\mypar{Evaluation of the compared methods.}
In Table \ref{tab:coverage_sample_level} we show the test set coverage achieved by the compared adapters at different levels of confidence. 
BayesAdapter is reliable at all confidence levels and achieves the highest coverage in all cases and across backbones, with a difference of 5-6\% over the second best approach. Observe also that earlier adapters are not generally reliable, specially when higher levels of confidence are requested.  

\begin{figure*}
\centering
    \begin{tabular}{cccc}
\includegraphics[width=0.19\textwidth]{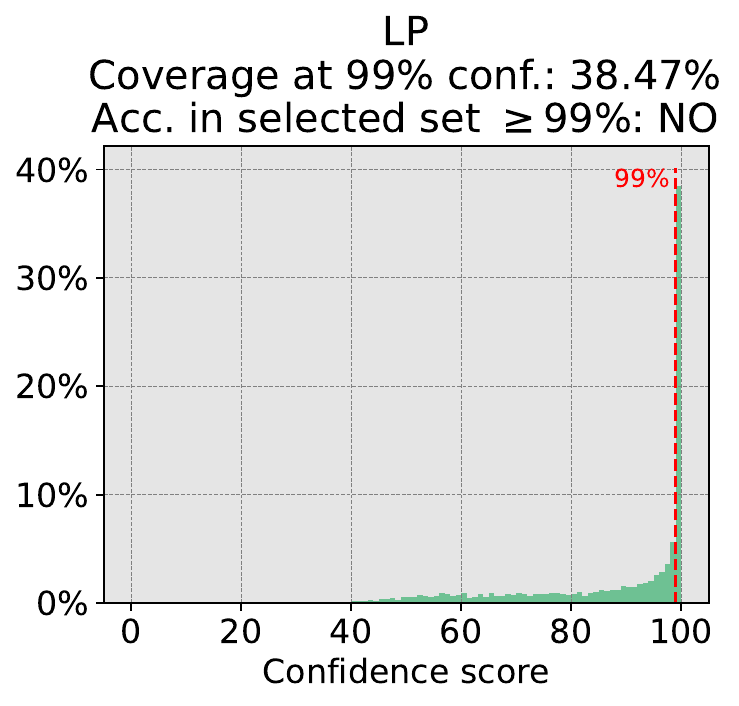} &
\includegraphics[width=0.19\textwidth]{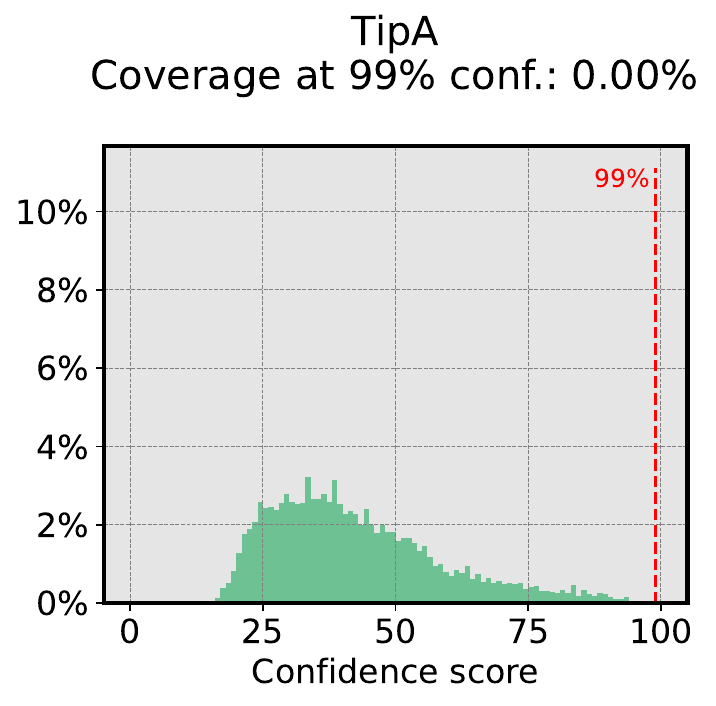} &
\includegraphics[width=0.19\textwidth]{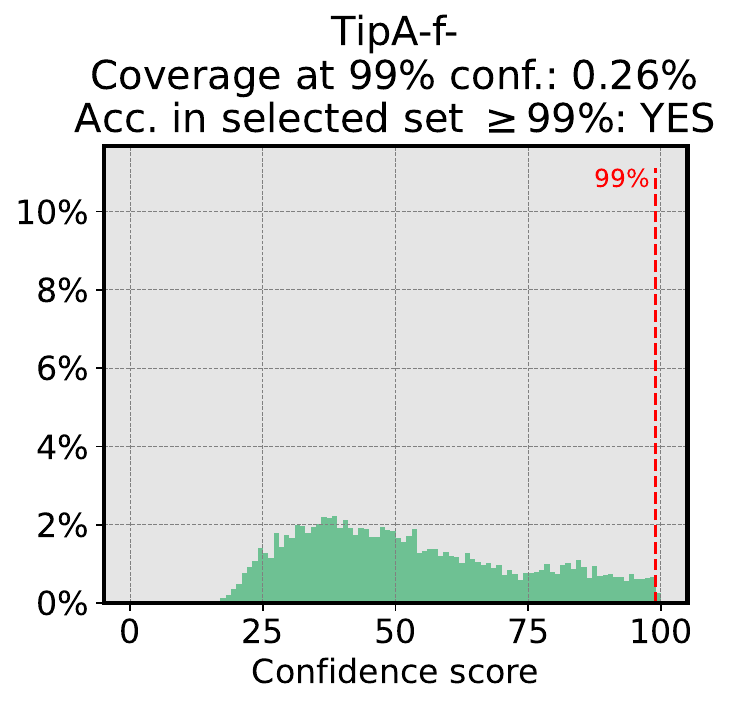}
& 
\includegraphics[width=0.19\textwidth]{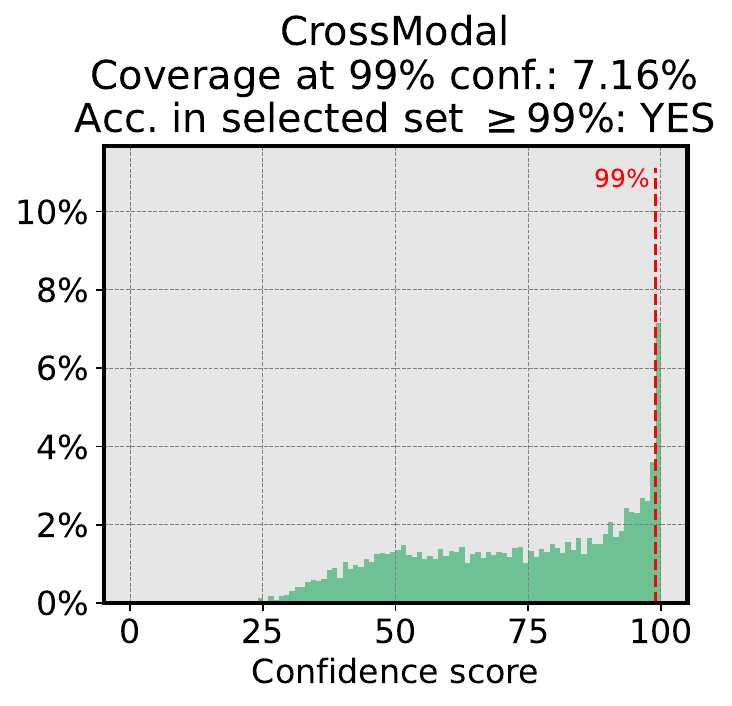}
\\
\noalign{\vskip -4pt}
\includegraphics[width=0.19\textwidth]{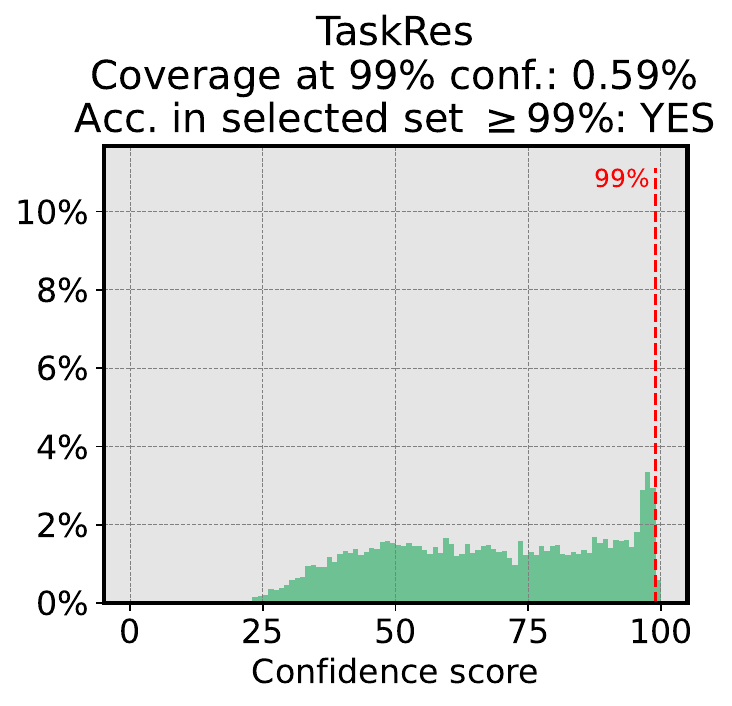}
& 
\includegraphics[width=0.19\textwidth]{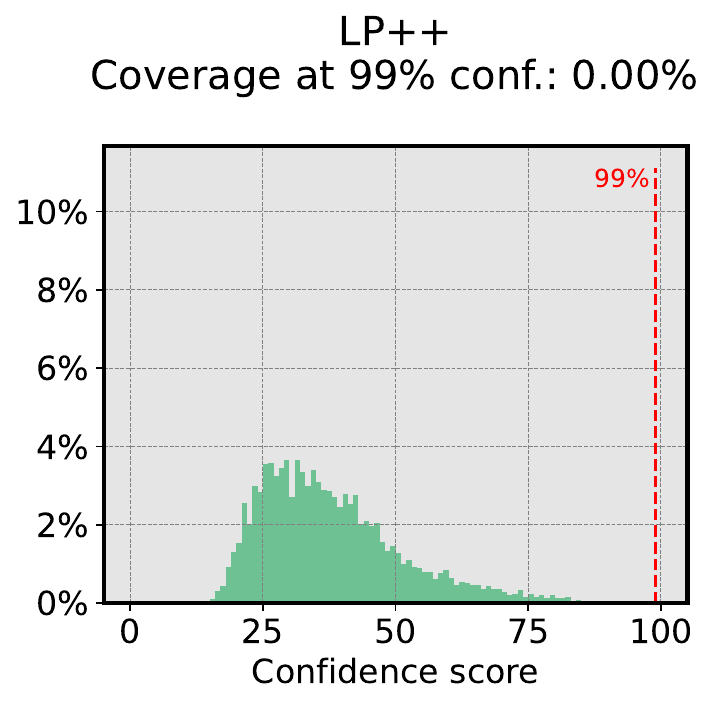}
& 
\includegraphics[width=0.19\textwidth]{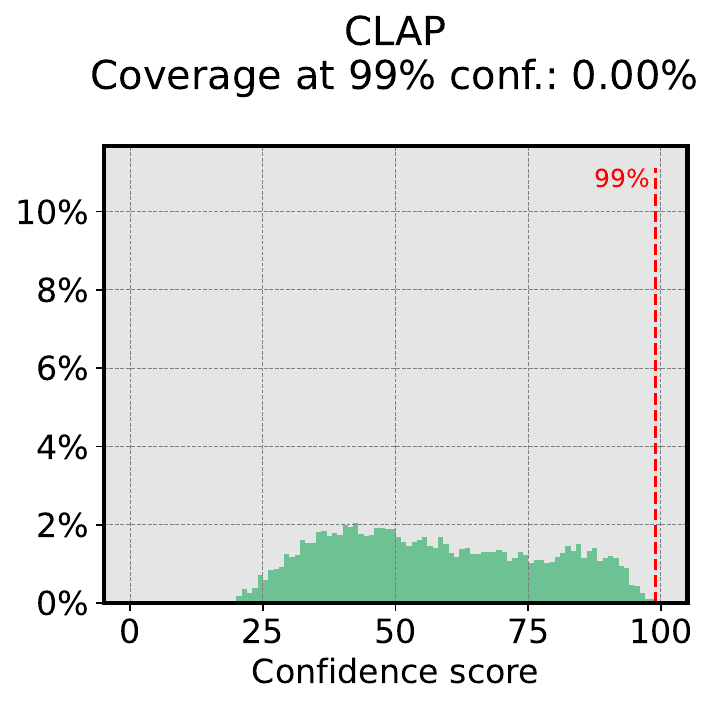}
& 
\includegraphics[width=0.19\textwidth]{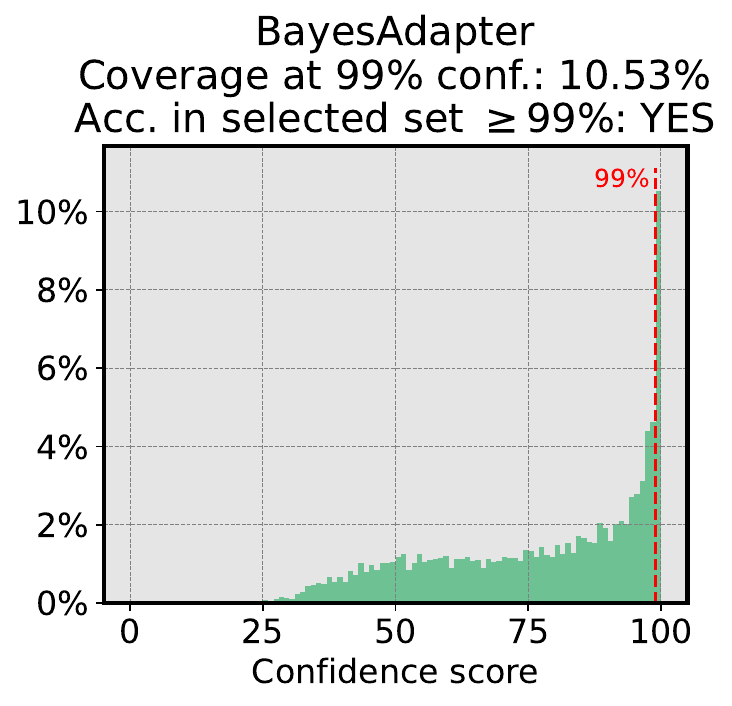}
\end{tabular}
\caption{\label{fig:eurosat}\textbf{Visualizing the over-conservative behavior of some baselines}, including the most recent ones CLAP and LP++. We show the histogram of the confidence score on the test samples (after adapting in a few-shot training data). 
Whereas CLAP, LP++ and TipA abstain from making predictions at 99\% confidence, BayesAdapter does cover 10.53\% of the test set, achieving accuracy above 99\%. TipA-f-, CrossModal and TaskRes also make reliable predictions at 99\% confidence, achieving coverage of 0.26\%, 7.16\% and 0.59\%, respectively. 
}
\end{figure*}

\mypar{Visualizing a paradigmatic example.}
There exist some paradigmatic examples that illustrate the too conservative behavior of many adapters.
Figure \ref{fig:eurosat} shows the histograms of confidence score for a run of the EuroSAT dataset with 4 shots. Notice that the top discriminative approaches CLAP and LP++ get a zero coverage at 99\% confidence, while BayesAdapter obtains 10.53\% (and the accuracy is above 99\% in the selected subset). 

\mypar{On the diversity of the selected samples.}
When deploying a system at high confidence level, another key aspect is the \emph{diversity of covered samples}. 
In practice, it is desirable that selected samples are as varied as possible (provided the method is reliable). 
Table \ref{tab:coverage_class_level} reports the coverage achieved by the different adapters in terms of classes, i.e., the percentage of classes that are represented in the selected subset (with at least one sample). 
BayesAdapter achieves the highest values at all confidence levels and across backbones, as in previous Table \ref{tab:coverage_sample_level}. 
However, the second-best method in previous table, CrossModal, is not the second-best here for many confidence levels.
This reveals a tendency to select samples from fewer classes. On the contrary, methods such as CLAP significantly improve when it comes to diversity of covered classes.

\begin{table*}
\footnotesize
    \centering
    \caption{\label{tab:coverage_class_level}
    \textbf{Comparison in terms of class-wise test set coverage at different confidence levels}.
    This measures the \% of classes that are represented among the samples with confidence above the threshold.
    {\color{red}\ding{55}} means that the method is not reliable at that confidence level (i.e. the accuracy in the selected subset is below the requested confidence).
    Best result in bold; second best underlined.}
    \begin{tabular}{lrrrrrrrrrr}
\toprule
 & \multicolumn{5}{c}{ResNet50} & \multicolumn{5}{c}{ViT-16} \\
 \cmidrule(lr){2-6}
 \cmidrule(lr){7-11}
 & \multicolumn{5}{c}{($\uparrow$) Class-wise test set coverage at confidence...} & \multicolumn{5}{c}{($\uparrow$) Class-wise test set coverage at confidence...} \\
 & 99\% & 95\% & 90\% & 85\% & 80\% & 99\% & 95\% & 90\% & 85\% & 80\% \\
\midrule
LP$_\text{ICML'21}$ & {\color{red} \ding{55}} & {\color{red} \ding{55}} & {\color{red} \ding{55}} & {\color{red} \ding{55}} & {\color{red} \ding{55}} & {\color{red} \ding{55}} & {\color{red} \ding{55}} & {\color{red} \ding{55}} & {\color{red} \ding{55}} & {\color{red} \ding{55}} \\
TipA$_\text{ECCV'22}$ & {\color{red} \ding{55}} & {\color{red} \ding{55}} & 80.589 & 84.937 & 88.111 & {\color{red} \ding{55}} & {\color{red} \ding{55}} & {\color{red} \ding{55}} & {\color{red} \ding{55}} & {\color{red} \ding{55}} \\
TipA-f-$_\text{ECCV'22}$ & {\color{red} \ding{55}} & \underline{75.395} & \underline{82.546} & 86.798 & 89.592 & {\color{red} \ding{55}} & {\color{red} \ding{55}} & {\color{red} \ding{55}} & {\color{red} \ding{55}} & {\color{red} \ding{55}} \\
CrossModal$_\text{CVPR'23}$ & \underline{48.916} & 72.110 & 81.212 & 85.883 & 89.193 & \underline{60.948} & \underline{80.550} & 87.788 & 91.398 & 93.747 \\
TaskRes$_\text{CVPR'23}$ & 43.699 & 69.283 & 79.353 & 84.210 & 87.442 & 56.756 & 78.498 & 86.071 & 89.719 & 91.963 \\
LP++$_\text{CVPR'24}$ & 23.523 & 49.551 & 62.304 & 69.830 & 75.161 & 42.684 & 65.726 & 76.177 & 81.848 & 85.703 \\
CLAP$_\text{CVPR'24}$ & 38.504 & 68.759 & 81.108 & \underline{87.160} & \underline{90.617} & 51.929 & 78.697 & \underline{88.336} & \underline{92.538} & \underline{94.852} \\
\gray BayesAdapter$_\text{Ours}$ & \gray\textbf{61.062} & \gray\textbf{81.935} & \gray\textbf{88.858} & \gray\textbf{91.740} & \gray\textbf{93.737} & \gray\textbf{72.115} & \gray\textbf{88.630} & \gray\textbf{93.410} & \gray\textbf{95.914} & \gray\textbf{97.513} \\
\bottomrule
\end{tabular}
\end{table*}

\subsubsection{Discriminative performance}

So far, we have verified that the probabilistic modeling of BayesAdapter leads to high quality uncertainty estimates. 
Here, we evaluate all the methods in terms of test accuracy. 
Table \ref{tab:acc_cal_all} shows that our BayesAdapter yields very competitive results. Indeed, the best approach, CLAP, outperforms BayesAdapter by $0.7\%$ and $0.55\%$ when employing ResNet-50 and ViT-16 as backbones, respectively.

\begin{figure*}[ht]
    \centering
    \begin{subfigure}{0.31\textwidth}  
        \includegraphics[width=\linewidth]{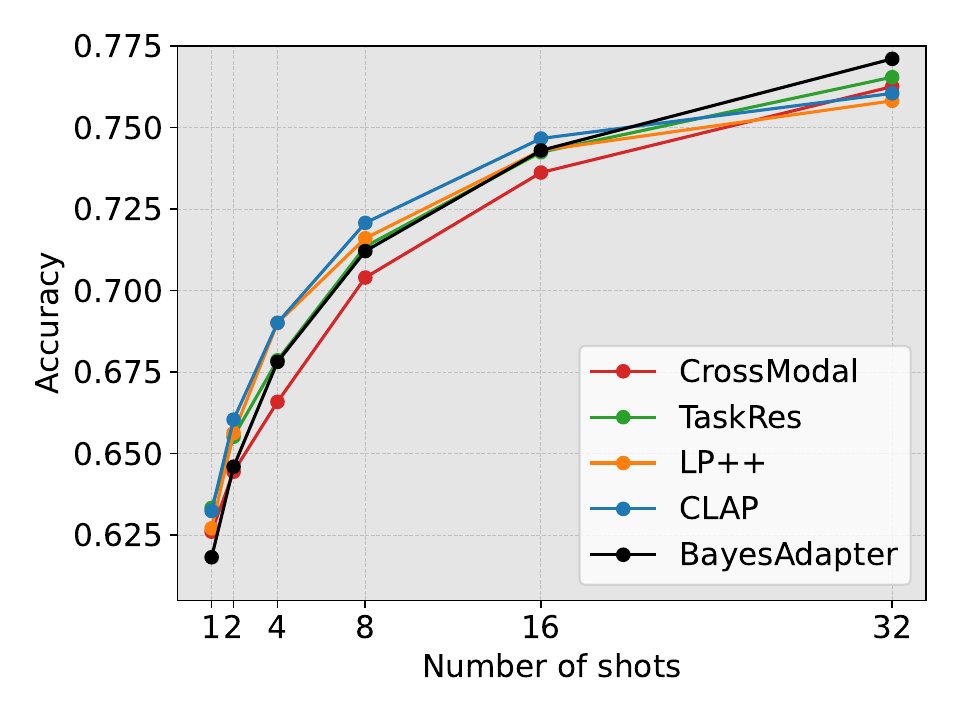}
    \end{subfigure}
    \hfill
    \begin{subfigure}{0.31\textwidth}
        \includegraphics[width=\linewidth]{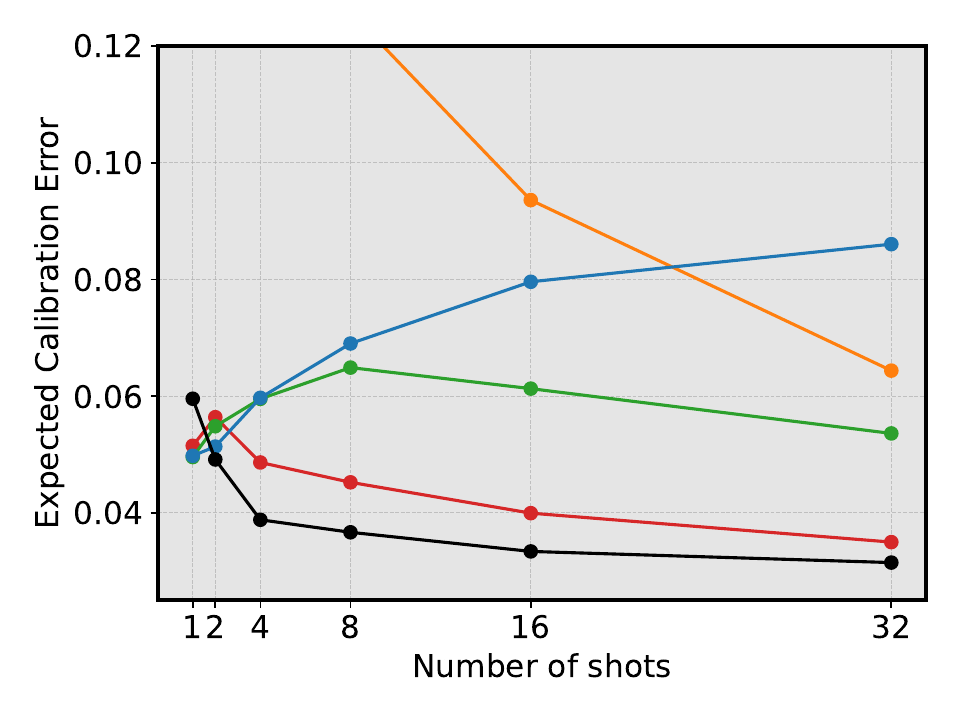}
    \end{subfigure}
    \hfill
    \begin{subfigure}{0.31\textwidth}
        \includegraphics[width=\linewidth]{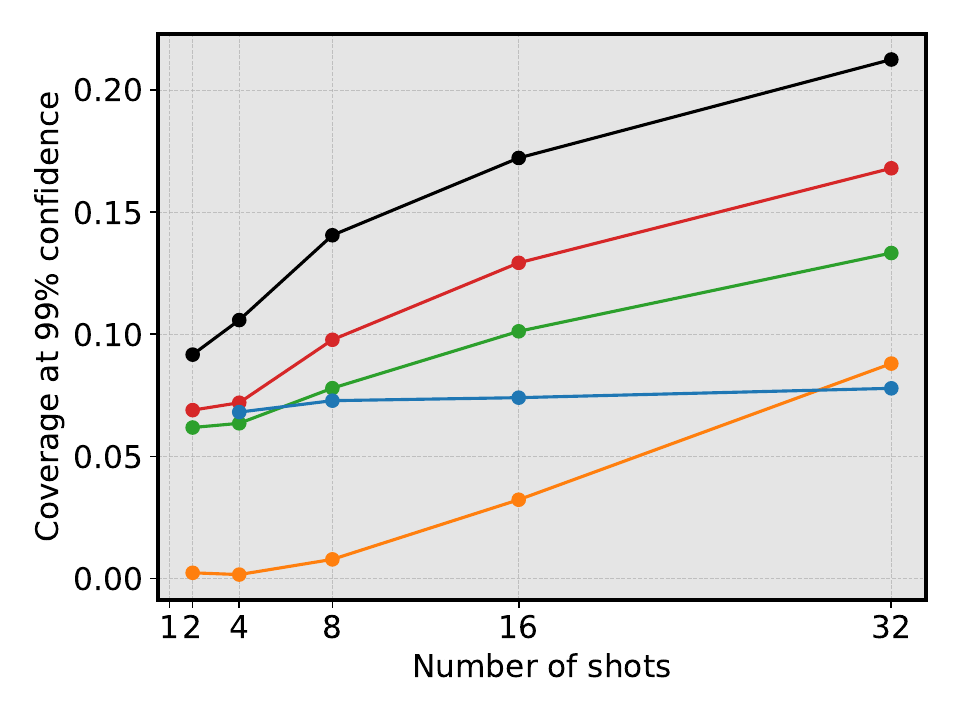}
    \end{subfigure}
    \caption{\label{fig:evol_shots}
    \textbf{Evolution of metrics with shots.}
    From left to right: accuracy, calibration, and coverage at 99\% confidence level. In the last one, the marker is missing if the method is not reliable at 99\% confidence level. 
    Whereas methods are similar in accuracy, more differences appear when evaluating the uncertainty estimates via calibration and selective classification. More details in the text.
    }
\end{figure*}

In principle \emph{one would expect the discriminative performance to improve} in the presence of a richer posterior distribution (instead of a single point estimate).
However, an important factor to consider here is the low number of shots utilized, which may not be sufficient to accurately estimate the additional parameters involved in BayesAdapter (posterior variance on top of posterior mean).
Indeed, when we analyze how the performance varies in terms of the number of shots, Figure \ref{fig:evol_shots}, we will see that BayesAdapter benefits from additional shots much more than CLAP does.
In fact, with 32 shots, BayesAdapter is the best-performing method in all the metrics, \emph{including test accuracy}.

\subsubsection{Additional results}

\mypar{Dependence on the number of shots.}
So far, we have analyzed accuracy, calibration and selective classification at high confidence by aggregating over all the number of shots available ($K\in \{1,2,4,8,16,32\}$).
Figure \ref{fig:evol_shots} shows how these metrics evolve w.r.t the number of shots used.
We include the five methods from 2023 onwards, since earlier approaches were not reliable at some confidence levels (recall Table \ref{tab:coverage_sample_level}), and their significantly lower performance make the rest of lines collapse in the plot.  
All numeric values are reported in the Appendix, Sec.~\ref{sec:app_full_results}.
The leftmost plot focuses on accuracy, where all methods improve with more shots, as theoretically expected. 
This improvement is particularly strong for BayesAdapter, which becomes the best performing one with 32 shots. 
This is explained by the need to estimate the posterior variance in addition to the mean, which benefits from some more shots.
Notice that differences in accuracy are smaller than in calibration (central plot) and coverage (rightmost plot).
Regarding calibration, Figure \ref{fig:evol_shots} reveals an interesting finding: the best method in test accuracy (CLAP, recall Table \ref{tab:acc_cal_all}) worsens its calibration with the number of shots.
This can be explained because the deterministic inference of CLAP lacks the richness needed to offset the zero-shot constraint, which becomes too restrictive as the number of shots increases. In contrast, the more flexible probabilistic approach of BayesAdapter effectively compensates for the prior distribution, resulting in a steady improvement as the number of shots increases.
Regarding selective classification at high confidence (rightmost plot),
the novel BayesAdapter consistently obtains the best results across shots with a significant margin.
Again, its deterministic counterpart, CLAP, struggles to clearly benefit from more shots. 
Notice that none of the methods are reliable at 99\% confidence when adapted with only 1 shot (that is why the lines start at $K=2$ shots; in fact, notice that CLAP becomes reliable only from $K=4$).

\mypar{Ablation study and training time for BayesAdapter.}
Table \ref{tab:ablation} ablates BayesAdapter under different values of its two main hyperparameters identified in Section \ref{sec:exp_setting}.
Similar to Table \ref{tab:acc_cal_all}, these results are aggregated over datasets, number of shots and seeds.
As is common in variational inference, the performance is similar across different values of $s_{\mathrm{MC}}$, since the mini-batch training across epochs allows for exploring the distribution space even for small values of samples. The impact of the prior standard deviation is larger. As theoretically expected, a very small prior prevents the method from leveraging the training shots, achieving poor accuracy and coverage.

Since $s_{\mathrm{MC}}=3$ Monte Carlo samples is enough during training (indeed it could be reduced to 1 as shown in Table \ref{tab:ablation}), in practice BayesAdapter does not imply a significant computational overhead w.r.t. the deterministic CLAP. This is empirically shown in Table \ref{tab:times}, see Sec.~\ref{sec:app_other_tab_figs} in the Appendix for a comparison with the rest of the adapters.

\begin{table}
\footnotesize
    \centering
    \caption{\textbf{Ablation study} for BayesAdapter. We validate different values for the two main hyperparameters identified in Section \ref{sec:exp_setting}: the number of samples used for MC integration during training and the prior standard deviation used for each class.\label{tab:ablation}}
    \begin{tabular}{cccc}
\toprule
 & Acc & ECE & Cov@99\% conf. \\
 \midrule
MC \# samples ($s_{\mathrm{MC}}$) &  &  &  \\
\midrule
1 & 69.443 & 4.130 & 9.253 \\
2 & 69.419 & 4.158 & 9.251 \\
3 & 69.437 & 4.168 & 9.213 \\
5 & 69.420 & 4.149 & 9.231 \\
10 & 69.420 & 4.173 & 9.254 \\
\midrule
prior std. dev. &  &  &  \\
\midrule
0.0001 & 58.693 & 4.554 & 3.062 \\
0.001 & 66.603 & 4.710 & 4.115 \\
0.01 & 69.437 & 4.168 & 9.213 \\
0.1 & 69.136 & 4.374 & 10.041 \\
1.0 & 68.747 & 4.791 & 12.043 \\
\bottomrule
\end{tabular}
\end{table}

\begin{table}
\footnotesize
    \centering
    \caption{\textbf{Comparison of training time} (in seconds) between the proposed BayesAdapter and its deterministic counterpart CLAP. This is the average time for adaptation in the Caltech101 dataset.\label{tab:times}}
    {
    \setlength{\tabcolsep}{5pt}
    \begin{tabular}{ccccccc}
         \toprule
         & \multicolumn{6}{c}{Number of shots} \\
         \cmidrule{2-7}
         &  1 & 2 & 4 & 8 & 16 & 32 \\
\midrule
CLAP$_\text{CVPR'24}$ & $2.64$ & $5.14$ & $4.27$ & $11.23$ & $20.94$ & $32.25$ \\
\gray BayesAdapter$_\text{Ours}$ & \gray$6.47$ & \gray$7.34$ & \gray$8.04$ & \gray$14.93$ & \gray$26.62$ & \gray$42.40$ \\
 \bottomrule
    \end{tabular}
    }
\end{table}

\section{Conclusions and limitations} 

Motivated by the increasing popularity of large vision-language models in real-world safety-critical scenarios, in this work we have evaluated current CLIP adapters beyond their discriminative performance, analyzing the quality of their uncertainty estimates in calibration and selective classification at high confidence. 
We have found that the best state-of-the-art (SOTA) adapters in discriminative performance do not always provide the best uncertainty estimation capabilities.
Then, we have shown that one of such SOTA adapters is just a particular case of a more general probabilistic formulation. 
By performing Bayesian inference in such method, which implies leveraging a posterior distribution instead of a single point estimate, we show that the quality of uncertainty estimates is remarkably improved. 
The main limitation of the proposed approach is that it struggles to fully exploit its richness in the presence of very few shots per-class, e.g., 1-shot. 
Yet, since the amount of training data is known before adaptation, in practice this could be a criteria to decide whether to use the deterministic or probabilistic formulation. 
We hope this works contributes towards a more comprehensive evaluation of CLIP adapters and related approaches, whose uncertainty should be systematically evaluated for a safe deployment.

\FloatBarrier

{
    \small
    \bibliographystyle{ieeenat_fullname}
    \bibliography{main}
}

\clearpage
\setcounter{page}{1}
\maketitlesupplementary

\appendix

\noindent
{
\normalsize

\section{Theoretical derivations}
\label{sec:app_theoretical_der}

\subsection{Proposition 1}

Here we justify step by step the equivalence stated in Proposition \ref{prop:1}, by which performing MAP inference in the novel BayesAdapter formulation is equivalent to minimizing the CLAP training objective. 

\begin{proposition_nn}
Given training data $(\bX,\bY)$, maximizing the (log) posterior probablity $\p(\bW|\bX,\bY)$ for the model in eqs.\eqref{eq:prob_model_prior}--\eqref{eq:prob_model_lik} is equivalent to minimizing the loss in eq.~\eqref{eq:CLAP_obj}.
\end{proposition_nn}

\noindent\textbf{Proof.}
Applying Bayes theorem, one has that
\begin{equation}\label{eq:bayes_theorem}
    \p(\bW|\bX,\bY) = \frac{\p(\bY|\bW,\bX)\p(\bW)}{\p(\bY|\bX)}.
\end{equation}
Thus, using that $\log$ is an increasing function and the fact that the denominator $\p(\bY|\bX)$ does not depend on $\bW$, we have:
\begin{multline}\label{eq:argmax}
\argmax_\bW \p(\bW|\bX,\bY) =
\argmax_\bW \left(\log\p(\bW|\bX,\bY)\right)= \\
=\argmax_\bW \left(\log\p(\bY|\bW,\bX)+\log\p(\bW)\right).    
\end{multline}
Let us expand each one of the two terms in the last expression. The first one is:
\begin{align}\label{eq:loglik_decomp}
    \log\p(\bY|\bW,\bX)&=\sum_{n=1}^N \log\p(\by_n|\bW,\bx_n)= \\ \nonumber
    &=\sum_{n=1}^N 
\log\left(\softmax(\psi_v(\bx_n)\cdot\bW)\cdot \by_n^\top\right) = \\ \nonumber
    & = \sum_{n=1}^N \log \hat y_{n,c(n)}
    = -\sum_{n=1}^N \mathcal{H}(\by_n,\hat\by_n),
\end{align}
where $\hat\by_n=\softmax(\psi_v(\bx_n)\cdot\bW)$ is a $C$-dimensional vector (recall that $C$ is the number of classes) and $c(n)$ is the class of the $n$-th instance (i.e. the location in which the one-hot encoding $\by_n$ is nonzero). 

Regarding the second term, taking into account the expression of a multivariante Gaussian distribution, we have:
\begin{align*}
    \log\p(\bW) &= \log\mathcal{N}(\bW|\bT,\bLambda) = 
    \\ \nonumber
    &=-0.5\cdot (\bW-\bT)\Lambda^{-1}(\bW-\bT)^\top+\mathrm{Ct.},
\end{align*}
where $\mathrm{Ct.}$ is a constant that do not depend on $\bW$, and notice that we are writing $\bW-\bT$ to denote the corresponding flattened row vector.
Remembering that $\bLambda=0.5\cdot\mathrm{diag}(\lambda_1^{-1},\dots,\lambda_1^{-1},\dots,\lambda_C^{-1},\dots,\lambda_C^{-1})$, we have:
\begin{align}\label{eq:prior_decomp}
    \log\p(\bW) &= -\sum_{c=1}^C \lambda_c\cdot ||\bw_c-\bt_c||^2.
\end{align}

Consdering the equality in eq.~\eqref{eq:argmax} and the expressions in eq.~\eqref{eq:loglik_decomp} and eq.~\eqref{eq:prior_decomp}, we obtain:
\begin{multline*}
    \argmax_\bW \p(\bW|\bX,\bY) =
    \\
    =\argmin_\bW \left( 
    \sum_{n=1}^N \mathcal{H}(\by_n,\hat\by_n)+
    \sum_{c=1}^C \lambda_c\cdot ||\bw_c-\bt_c||^2
    \right),
\end{multline*}
which is the stated equivalence.

\subsection{Equivalence between KL minimization and negative ELBO minimization}

To derive the BayesAdapter training goal, recall eq.~\eqref{eq:ours_obj}, we stated that \emph{minimizing the KL
divergence between the approximate parametric posterior and the true posterior is equivalent to minimizing the negative Evidence Lower Bound (ELBO)}. This is a well-known property in variational inference, and we derive it here step-by-step for our model.

\begin{claim_nn}
Consider the the probabilistic model stated in eqs.~\eqref{eq:prob_model_prior}--\eqref{eq:prob_model_lik}, training data $(\bX,\bY)$, and a parametric distribution $\q_\alpha(\bW)$. Then:
\begin{multline*}
    \min_\alpha
    \mathrm{KL}\left(\q_\alpha(\bW)||\p(\bW|\bX,\bY)\right)=\\
    =\min_{\alpha}\left[\sum_{n=1}^N\left( \Ebb_{\q_\alpha(\bW)}\cH(\by_n, \hat\by_n)\right)\!+\!\mathrm{KL}(\q_\alpha(\bW)||\p(\bW))\right].
\end{multline*}
\end{claim_nn}

\noindent\textbf{Proof.}
The definition of KL divergence is:
\begin{multline*}
\mathrm{KL}=\mathrm{KL}\left(\q_\alpha(\bW)||\p(\bW|\bX,\bY)\right)=\\
=\int \q_\alpha(\bW) \log\frac{\q_\alpha(\bW)}{\p(\bW|\bX,\bY)} \mathrm{d}\bW.
\end{multline*}
Notice that the denominator can be rewritten using Bayes theorem, recall eq.~\eqref{eq:bayes_theorem}, and rearranging terms we have:
{\footnotesize
\begin{multline*}
    \int \q_\alpha(\bW) \left(
\log\p(\bY|\bX)-\log \p(\bY|\bW,\bX)+\log\frac{\q_\alpha(\bW)}{\p(\bW)}
    \right) \mathrm{d}\bW.
\end{multline*}
}
Taking into account that $\int \q_\alpha(\bW)\mathrm{d}\bW = 1$ (because it is a probability distribution) and considering the expression for $\log\p(\bY|\bW,\bX)$ obtained in eq.~\eqref{eq:loglik_decomp}, we have:
\begin{multline*}
    \mathrm{KL}=
    \log\p(\bY|\bX)+
    \sum_{n=1}^N\left( \Ebb_{\q_\alpha(\bW)}\cH(\by_n, \hat\by_n)\right)+\\
    +\mathrm{KL}(\q_\alpha(\bW)||\p(\bW)).
\end{multline*}
Since $\log\p(\bY|\bX)$ does not depend on $\alpha$, minimizing w.r.t. $\alpha$ in the last equation yields the sought equivalence.

\section{Other figures and tables}\label{sec:app_other_tab_figs}

\noindent Calibration plots for all the methods: Figure \ref{fig:calib_plots_full}

\noindent Effect of the technique proposed in \cite{murugesan2024robust}: Table \ref{tab:sals}

\noindent Training times for all the methods: Table \ref{tab:training_times_full}

\section{Full numerical results for the reported metrics}\label{sec:app_full_results}

Using \textbf{ResNet-50} backbone:
\begin{itemize}
    \item Test accuracy: Table \ref{tab:full_acc_rn50}
    \item ECE: Table \ref{tab:full_ece_rn50}
    \item AECE: Table \ref{tab:full_aece_rn50}
    \item Selective classification at 99\% confidence: Table \ref{tab:full_SC_rn50_99}
    \item Selective classification at 95\% confidence: Table \ref{tab:full_SC_rn50_95}
    \item Selective classification at 90\% confidence: Table \ref{tab:full_SC_rn50_90}
    \item Selective classification at 85\% confidence: Table \ref{tab:full_SC_rn50_85}
    \item Selective classification at 80\% confidence: Table \ref{tab:full_SC_rn50_80}
\end{itemize}

\vspace{10mm}

\noindent Using \textbf{ViT-B/16} backbone:
\begin{itemize}
    \item Test accuracy: Table \ref{tab:full_acc_vit16}
    \item ECE: Table \ref{tab:full_ece_vit16}
    \item AECE: Table \ref{tab:full_aece_vit16}
    \item Selective classification at 99\% confidence: Table \ref{tab:full_SC_vit16_99}
    \item Selective classification at 95\% confidence: Table \ref{tab:full_SC_vit16_95}
    \item Selective classification at 90\% confidence: Table \ref{tab:full_SC_vit16_90}
    \item Selective classification at 85\% confidence: Table \ref{tab:full_SC_vit16_85}
    \item Selective classification at 80\% confidence: Table \ref{tab:full_SC_vit16_80}
\end{itemize}

\FloatBarrier

\begin{figure*}[ht]
    \centering
    \begin{subfigure}{0.23\textwidth}  
        \includegraphics[width=\linewidth]{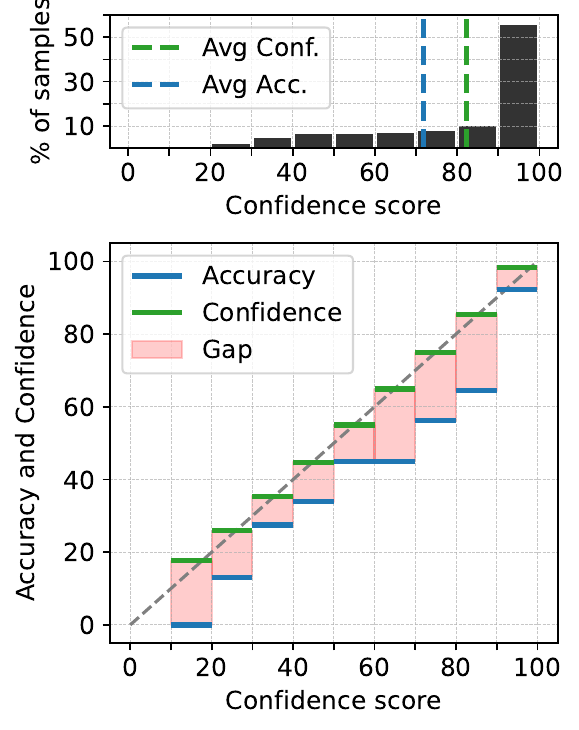}
        \caption{LP}
    \end{subfigure}
    \hfill
    \begin{subfigure}{0.23\textwidth}
        \includegraphics[width=\linewidth]{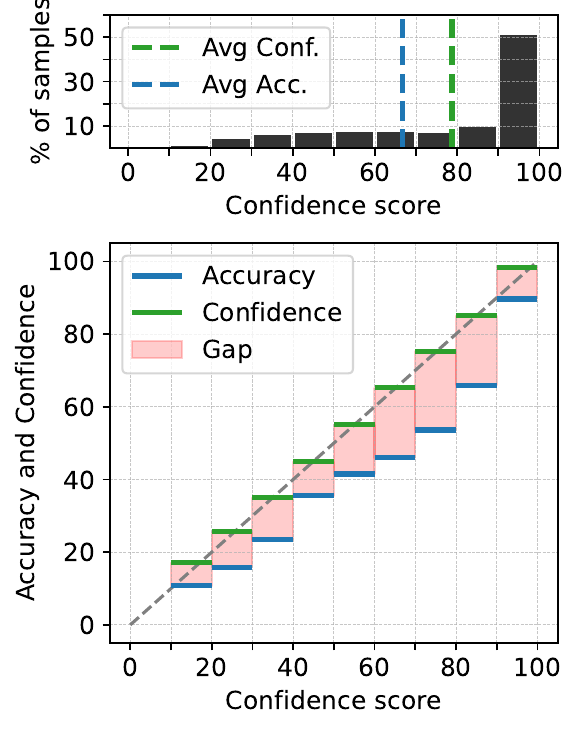}
        \caption{TipA}
    \end{subfigure}
    \hfill
    \begin{subfigure}{0.23\textwidth}
        \includegraphics[width=\linewidth]{figures/TipA-f-.pdf}
        \caption{TipA-f-}
    \end{subfigure}
    \hfill
    \begin{subfigure}{0.23\textwidth}
        \includegraphics[width=\linewidth]{figures/CrossModal.pdf}
        \caption{CrossModal}
    \end{subfigure}

    \vspace{1mm}  
    \begin{subfigure}{0.23\textwidth}
        \includegraphics[width=\linewidth]{figures/TR.pdf}
        \caption{TaskRes}
    \end{subfigure}
    \hfill
    \begin{subfigure}{0.23\textwidth}
        \includegraphics[width=\linewidth]{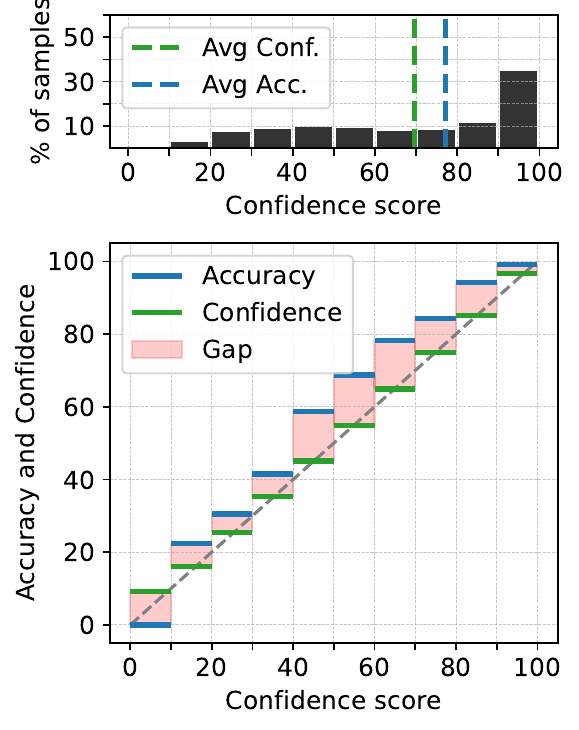}
        \caption{LP++}
    \end{subfigure}
    \hfill
    \begin{subfigure}{0.23\textwidth}
        \includegraphics[width=\linewidth]{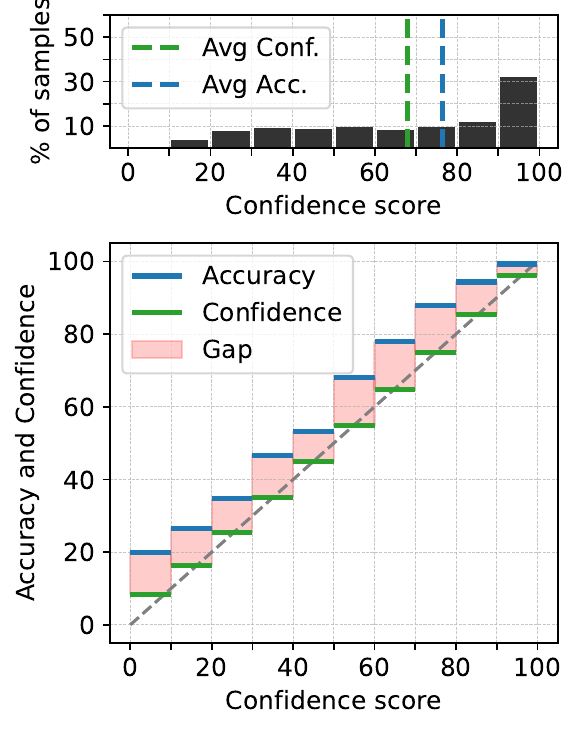}
        \caption{CLAP}
    \end{subfigure}
    \hfill
    \begin{subfigure}{0.23\textwidth}
        \includegraphics[width=\linewidth]{figures/BayesAdapter.pdf}
        \caption{BayesAdapter (ours)}
    \end{subfigure}
    \caption{\textbf{Calibration plots} for all the compared methods (ResNet50 backbone).
    In each case, the lower subplot depicts the accuracy and average confidence for samples in each one of the ten bins (from 0\% to 100\% of confidence score by steps of 10\%). Ideally, the gap between them should be zero. The upper plot shows the proportion of samples in each bin, along with the average confidence and accuracy in the whole test set. 
    \label{fig:calib_plots_full}
    }
\end{figure*}

\begin{table*}
    \centering
    \caption{\textbf{Evaluation of SALS} (sample-adaptive logit scaling), the logit normalization technique introduced in \cite{murugesan2024robust}.
    This technique was introduced to improve calibration in the domain shift scenario and only tested when adapting on ImageNet. 
    When evaluated on our more extensive setting, we observe that it generally degrades the calibration, staying consistently below the performance of the novel BayesAdapter.
    These results are the average over all the datasets, number of shots and seeds, and we are using the ResNet-50 backbone.
    \label{tab:sals}}
\begin{tabular}{lcccc}
\toprule
 & \multicolumn{2}{c}{ECE} & \multicolumn{2}{c}{AECE} \\
 \cmidrule(lr){2-3} 
 \cmidrule(lr){4-5}
 & No postprocessing & SALS postprocessing & No postprocessing & SALS postprocessing \\
\midrule
LP$_\text{ICML'21}$ & 24.307 & 24.873 & 24.297 & 24.860 \\
TipA$_\text{ECCV'22}$ & 7.772 & 6.200 & 7.740 & 6.152 \\
TipA-f-$_\text{ECCV'22}$ & 5.865 & 9.042 & 5.799 & 9.026 \\
CrossModal$_\text{CVPR'23}$ & 4.611 & 9.714 & 4.601 & 9.693 \\
TaskRes$_\text{CVPR'23}$ & 5.729 & 10.453 & 5.735 & 10.439 \\
LP++$_\text{CVPR'24}$ & 13.812 & 9.870 & 13.821 & 9.850 \\
CLAP$_\text{CVPR'24}$ & 6.592 & 10.331 & 6.572 & 10.316 \\
\midrule
BayesAdapter$_\text{Ours}$ & \textbf{4.151} & - & \textbf{4.098} & - \\
\bottomrule
\end{tabular}
\end{table*}

\begin{table*}
\centering
\caption{\textbf{Comparison of training time} (in seconds) between all the compared methods.
We show the average and standard error over three independent runs of adaptation in the Caltech101 dataset.
In general, all CLIP adapters provide a very fast adaptation to downstream tasks following their parameter-efficient design.
\label{tab:training_times_full}}
\begin{tabular}{lcccccc}
\toprule
& \multicolumn{6}{c}{Number of shots per class} \\
\cmidrule{2-7}
 & 1 & 2 & 4 & 8 & 16 & 32 \\
\midrule
LP$_\text{ICML'21}$ & $2.60_{\pm 0.28}$ & $2.66_{\pm 0.30}$ & $8.64_{\pm 0.78}$ & $11.48_{\pm 0.95}$ & $18.82_{\pm 1.85}$ & $51.85_{\pm 5.20}$ \\
TipA$_\text{ECCV'22}$ & $0.71_{\pm 0.01}$ & $1.10_{\pm 0.30}$ & $0.50_{\pm 0.01}$ & $0.79_{\pm 0.04}$ & $1.93_{\pm 0.07}$ & $0.99_{\pm 0.18}$ \\
TipA-f-$_\text{ECCV'22}$ & $2.49_{\pm 0.09}$ & $2.48_{\pm 0.23}$ & $2.99_{\pm 0.21}$ & $10.63_{\pm 1.10}$ & $24.92_{\pm 2.44}$ & $80.11_{\pm 7.89}$ \\
CrossModal$_\text{CVPR'23}$ & $2.94_{\pm 0.33}$ & $3.52_{\pm 0.27}$ & $12.66_{\pm 1.01}$ & $16.30_{\pm 1.70}$ & $30.69_{\pm 3.16}$ & $69.70_{\pm 6.90}$ \\
TaskRes$_\text{CVPR'23}$ & $2.50_{\pm 0.37}$ & $3.41_{\pm 0.46}$ & $5.60_{\pm 0.52}$ & $10.38_{\pm 1.01}$ & $18.35_{\pm 1.73}$ & $51.37_{\pm 5.00}$ \\
LP++$_\text{CVPR'24}$ & $1.12_{\pm 0.34}$ & $1.76_{\pm 0.03}$ & $2.29_{\pm 0.23}$ & $7.47_{\pm 0.73}$ & $16.24_{\pm 1.78}$ & $28.04_{\pm 2.95}$ \\
CLAP$_\text{CVPR'24}$ & $2.64_{\pm 0.34}$ & $5.14_{\pm 0.53}$ & $4.27_{\pm 0.47}$ & $11.23_{\pm 1.16}$ & $20.94_{\pm 2.24}$ & $32.25_{\pm 3.20}$ \\
BayesAdapter$_\text{Ours}$ & $6.47_{\pm 0.74}$ & $7.34_{\pm 0.72}$ & $8.04_{\pm 0.79}$ & $14.93_{\pm 1.53}$ & $26.62_{\pm 2.68}$ & $42.40_{\pm 4.39}$ \\
\bottomrule
\end{tabular}
\end{table*}

\begin{table*}[ht]
{\tiny
\setlength{\tabcolsep}{2.5pt}
    \centering
    \begin{tabular}{lllllllllllll}
\toprule
 & dataset & Caltech101 & DescribTextures & EuroSAT & FGVCAircraft & Food101 & ImageNet & OxfordFlowers & OxfordPets & SUN397 & StanfordCars & UCF101 \\
num shots & method &  &  &  &  &  &  &  &  &  &  &  \\
\midrule
\multirow[t]{8}{*}{1} & LP$_\text{ICML'21}$ & $57.96_{\pm 2.12}$ & $28.37_{\pm 0.62}$ & $42.93_{\pm 3.93}$ & $12.06_{\pm 0.17}$ & $26.09_{\pm 0.15}$ & $17.88_{\pm 0.03}$ & $52.20_{\pm 1.72}$ & $28.19_{\pm 1.69}$ & $26.69_{\pm 0.27}$ & $19.57_{\pm 0.72}$ & $33.53_{\pm 2.04}$ \\
 & TipA$_\text{ECCV'22}$ & $84.95_{\pm 0.09}$ & $44.27_{\pm 0.21}$ & $42.77_{\pm 0.65}$ & $17.76_{\pm 0.12}$ & $75.12_{\pm 0.04}$ & $60.46_{\pm 0.02}$ & $67.61_{\pm 0.21}$ & $83.57_{\pm 0.15}$ & $58.74_{\pm 0.06}$ & $57.55_{\pm 0.10}$ & $59.58_{\pm 0.28}$ \\
 & TipA-f-$_\text{ECCV'22}$ & $86.13_{\pm 0.42}$ & $45.29_{\pm 0.28}$ & $52.00_{\pm 0.52}$ & $18.61_{\pm 0.34}$ & $75.46_{\pm 0.02}$ & $60.60_{\pm 0.02}$ & $69.20_{\pm 0.16}$ & $85.05_{\pm 0.58}$ & $59.10_{\pm 0.04}$ & $58.16_{\pm 0.16}$ & $61.34_{\pm 0.30}$ \\
 & CrossModal$_\text{CVPR'23}$ & $88.07_{\pm 0.37}$ & $47.85_{\pm 1.06}$ & $57.88_{\pm 3.03}$ & $20.44_{\pm 0.32}$ & $72.80_{\pm 0.20}$ & $57.89_{\pm 0.04}$ & $78.66_{\pm 0.50}$ & $81.43_{\pm 1.16}$ & $61.57_{\pm 0.15}$ & $57.63_{\pm 0.38}$ & $64.45_{\pm 0.77}$ \\
 & TaskRes$_\text{CVPR'23}$ & $88.52_{\pm 0.29}$ & $49.00_{\pm 1.16}$ & $58.98_{\pm 2.72}$ & $20.46_{\pm 0.12}$ & $74.07_{\pm 0.04}$ & $60.69_{\pm 0.04}$ & $76.18_{\pm 0.35}$ & $83.16_{\pm 1.10}$ & $61.31_{\pm 0.13}$ & $59.31_{\pm 0.13}$ & $65.02_{\pm 0.43}$ \\
 & LP++$_\text{CVPR'24}$ & $88.21_{\pm 0.48}$ & $46.43_{\pm 0.67}$ & $50.50_{\pm 1.11}$ & $20.13_{\pm 0.21}$ & $76.39_{\pm 1.12}$ & $60.96_{\pm 0.12}$ & $80.13_{\pm 0.12}$ & $83.51_{\pm 2.04}$ & $62.29_{\pm 0.13}$ & $57.29_{\pm 0.14}$ & $63.94_{\pm 0.57}$ \\
\bottomrule
\end{tabular}
}
    \caption{Accuracy with ResNet-50 backbone: full numerical results.}
    \label{tab:full_acc_rn50}
\end{table*}

\begin{table*}[ht]
{\tiny
\setlength{\tabcolsep}{2.5pt}
    \centering
   \begin{tabular}{lllllllllllll}
\toprule
 & dataset & Caltech101 & DescribTextures & EuroSAT & FGVCAircraft & Food101 & ImageNet & OxfordFlowers & OxfordPets & SUN397 & StanfordCars & UCF101 \\
num shots & method &  &  &  &  &  &  &  &  &  &  &  \\
\midrule
\multirow[t]{8}{*}{1} & LP$_\text{ICML'21}$ & $30.71_{\pm 1.50}$ & $56.38_{\pm 0.76}$ & $43.73_{\pm 1.33}$ & $59.82_{\pm 1.13}$ & $52.54_{\pm 0.07}$ & $46.66_{\pm 0.17}$ & $27.47_{\pm 1.07}$ & $51.33_{\pm 1.26}$ & $49.67_{\pm 0.19}$ & $51.81_{\pm 1.44}$ & $48.99_{\pm 2.24}$ \\
 & TipA$_\text{ECCV'22}$ & $2.44_{\pm 0.15}$ & $4.42_{\pm 0.26}$ & $6.53_{\pm 0.27}$ & $4.11_{\pm 0.14}$ & $2.32_{\pm 0.01}$ & $1.99_{\pm 0.07}$ & $3.86_{\pm 0.26}$ & $5.71_{\pm 0.19}$ & $1.85_{\pm 0.02}$ & $4.49_{\pm 0.15}$ & $2.54_{\pm 0.46}$ \\
 & TipA-f-$_\text{ECCV'22}$ & $2.74_{\pm 0.14}$ & $5.54_{\pm 0.36}$ & $15.96_{\pm 0.91}$ & $3.55_{\pm 0.43}$ & $2.60_{\pm 0.04}$ & $1.92_{\pm 0.06}$ & $5.11_{\pm 0.12}$ & $6.66_{\pm 0.51}$ & $1.99_{\pm 0.03}$ & $4.86_{\pm 0.22}$ & $3.41_{\pm 0.27}$ \\
 & CrossModal$_\text{CVPR'23}$ & $2.22_{\pm 0.33}$ & $5.16_{\pm 0.48}$ & $11.54_{\pm 2.78}$ & $9.32_{\pm 0.19}$ & $1.46_{\pm 0.12}$ & $4.22_{\pm 0.04}$ & $10.73_{\pm 0.46}$ & $2.37_{\pm 0.56}$ & $4.40_{\pm 0.10}$ & $1.88_{\pm 0.34}$ & $3.33_{\pm 0.74}$ \\
 & TaskRes$_\text{CVPR'23}$ & $3.47_{\pm 0.31}$ & $3.49_{\pm 0.51}$ & $5.76_{\pm 2.50}$ & $4.80_{\pm 0.40}$ & $2.63_{\pm 0.13}$ & $1.77_{\pm 0.04}$ & $11.71_{\pm 0.43}$ & $4.44_{\pm 0.62}$ & $5.42_{\pm 0.15}$ & $5.30_{\pm 0.14}$ & $5.73_{\pm 0.30}$ \\
 & LP++$_\text{CVPR'24}$ & $14.47_{\pm 0.39}$ & $17.37_{\pm 1.55}$ & $17.42_{\pm 1.89}$ & $5.23_{\pm 1.75}$ & $19.62_{\pm 2.08}$ & $25.84_{\pm 0.26}$ & $21.04_{\pm 1.83}$ & $14.08_{\pm 3.57}$ & $27.81_{\pm 1.28}$ & $16.15_{\pm 1.82}$ & $28.17_{\pm 1.64}$ \\
\bottomrule
\end{tabular}
}
\caption{ECE with ResNet-50 backbone: full numerical results.}
    \label{tab:full_ece_rn50}
\end{table*}

\begin{table*}[ht]
{\tiny
\setlength{\tabcolsep}{2.5pt}
    \centering
\begin{tabular}{lllllllllllll}
\toprule
 & dataset & Caltech101 & DescribTextures & EuroSAT & FGVCAircraft & Food101 & ImageNet & OxfordFlowers & OxfordPets & SUN397 & StanfordCars & UCF101 \\
num shots & method &  &  &  &  &  &  &  &  &  &  &  \\
\midrule
\multirow[t]{8}{*}{1} & LP$_\text{ICML'21}$ & $30.70_{\pm 1.50}$ & $56.37_{\pm 0.76}$ & $43.73_{\pm 1.33}$ & $59.82_{\pm 1.13}$ & $52.54_{\pm 0.07}$ & $46.66_{\pm 0.17}$ & $27.46_{\pm 1.07}$ & $51.32_{\pm 1.26}$ & $49.67_{\pm 0.19}$ & $51.81_{\pm 1.44}$ & $48.99_{\pm 2.24}$ \\
 & TipA$_\text{ECCV'22}$ & $2.10_{\pm 0.24}$ & $4.59_{\pm 0.25}$ & $6.42_{\pm 0.27}$ & $4.11_{\pm 0.14}$ & $2.31_{\pm 0.01}$ & $1.96_{\pm 0.05}$ & $3.81_{\pm 0.11}$ & $5.58_{\pm 0.18}$ & $1.83_{\pm 0.05}$ & $4.43_{\pm 0.15}$ & $2.19_{\pm 0.43}$ \\
 & TipA-f-$_\text{ECCV'22}$ & $2.55_{\pm 0.18}$ & $5.35_{\pm 0.24}$ & $15.84_{\pm 0.98}$ & $3.56_{\pm 0.37}$ & $2.60_{\pm 0.04}$ & $1.90_{\pm 0.05}$ & $5.12_{\pm 0.08}$ & $6.65_{\pm 0.50}$ & $2.04_{\pm 0.06}$ & $4.83_{\pm 0.22}$ & $3.34_{\pm 0.51}$ \\
 & CrossModal$_\text{CVPR'23}$ & $2.31_{\pm 0.30}$ & $5.52_{\pm 0.37}$ & $11.54_{\pm 2.78}$ & $9.32_{\pm 0.19}$ & $1.42_{\pm 0.10}$ & $4.23_{\pm 0.05}$ & $10.68_{\pm 0.48}$ & $2.42_{\pm 0.53}$ & $4.39_{\pm 0.10}$ & $2.05_{\pm 0.31}$ & $3.31_{\pm 0.65}$ \\
 & TaskRes$_\text{CVPR'23}$ & $3.37_{\pm 0.27}$ & $3.87_{\pm 0.63}$ & $5.90_{\pm 2.42}$ & $4.70_{\pm 0.39}$ & $2.60_{\pm 0.13}$ & $1.76_{\pm 0.03}$ & $11.70_{\pm 0.42}$ & $4.23_{\pm 0.76}$ & $5.42_{\pm 0.15}$ & $5.29_{\pm 0.15}$ & $5.72_{\pm 0.29}$ \\
 & LP++$_\text{CVPR'24}$ & $14.47_{\pm 0.39}$ & $17.38_{\pm 1.55}$ & $17.33_{\pm 1.95}$ & $5.33_{\pm 1.50}$ & $19.62_{\pm 2.08}$ & $25.84_{\pm 0.26}$ & $21.02_{\pm 1.82}$ & $14.08_{\pm 3.58}$ & $27.81_{\pm 1.28}$ & $16.15_{\pm 1.82}$ & $28.17_{\pm 1.63}$ \\
\bottomrule
\end{tabular}

}
    \caption{AECE with ResNet-50 backbone: full numerical results.}
    \label{tab:full_aece_rn50}
\end{table*}

\begin{table*}[ht]
{\tiny
\setlength{\tabcolsep}{2.5pt}
    \centering
\begin{tabular}{lllllllllllll}
\toprule
 & dataset & Caltech101 & DescribTextures & EuroSAT & FGVCAircraft & Food101 & ImageNet & OxfordFlowers & OxfordPets & SUN397 & StanfordCars & UCF101 \\
num shots & method &  &  &  &  &  &  &  &  &  &  &  \\
\midrule
\multirow[t]{8}{*}{1} & LP$_\text{ICML'21}$ & {\color{red} \ding{55}} & {\color{red} \ding{55}} & {\color{red} \ding{55}} & {\color{red} \ding{55}} & {\color{red} \ding{55}} & {\color{red} \ding{55}} & {\color{red} \ding{55}} & {\color{red} \ding{55}} & {\color{red} \ding{55}} & {\color{red} \ding{55}} & {\color{red} \ding{55}} \\
 & TipA$_\text{ECCV'22}$ & $24.71_{\pm 0.18}$ & $0.85_{\pm 0.05}$ & $0.00_{\pm 0.00}$ & $0.23_{\pm 0.04}$ & $6.97_{\pm 0.09}$ & {\color{red} \ding{55}} & $1.89_{\pm 0.01}$ & $6.05_{\pm 0.12}$ & $1.26_{\pm 0.01}$ & $1.75_{\pm 0.02}$ & $2.58_{\pm 0.09}$ \\
 & TipA-f-$_\text{ECCV'22}$ & $25.46_{\pm 0.13}$ & $0.73_{\pm 0.05}$ & $0.00_{\pm 0.00}$ & $0.29_{\pm 0.09}$ & $7.15_{\pm 0.07}$ & {\color{red} \ding{55}} & $2.40_{\pm 0.08}$ & $6.82_{\pm 0.12}$ & {\color{red} \ding{55}} & $1.83_{\pm 0.03}$ & $2.74_{\pm 0.11}$ \\
 & CrossModal$_\text{CVPR'23}$ & $32.49_{\pm 0.26}$ & $2.21_{\pm 0.15}$ & {\color{red} \ding{55}} & $0.64_{\pm 0.15}$ & $10.08_{\pm 0.26}$ & {\color{red} \ding{55}} & $5.01_{\pm 0.49}$ & $12.22_{\pm 0.11}$ & {\color{red} \ding{55}} & $2.97_{\pm 0.04}$ & $4.13_{\pm 0.29}$ \\
 & TaskRes$_\text{CVPR'23}$ & $29.80_{\pm 0.43}$ & $1.62_{\pm 0.14}$ & {\color{red} \ding{55}} & $0.50_{\pm 0.10}$ & $9.85_{\pm 0.23}$ & {\color{red} \ding{55}} & $3.87_{\pm 0.43}$ & $11.35_{\pm 0.09}$ & $1.42_{\pm 0.05}$ & $2.61_{\pm 0.04}$ & $3.42_{\pm 0.23}$ \\
 & LP++$_\text{CVPR'24}$ & $0.31_{\pm 0.14}$ & $0.00_{\pm 0.00}$ & {\color{red} \ding{55}} & $0.00_{\pm 0.00}$ & $0.05_{\pm 0.03}$ & $0.01_{\pm 0.00}$ & $0.11_{\pm 0.06}$ & $0.40_{\pm 0.28}$ & $0.00_{\pm 0.00}$ & $0.05_{\pm 0.03}$ & $0.00_{\pm 0.00}$ \\
\bottomrule
\end{tabular}
}
    \caption{Selective classification at 99\% confidence, using ResNet-50 backbone. We show the test set coverage provided the accuracy on the selected test set is greater than or equal to the requested confidence. Otherwise, the method is not reliable at such confidence level and we report {\color{red} \ding{55}}.}
    \label{tab:full_SC_rn50_99}
\end{table*}

\begin{table*}[ht]
{\tiny
\setlength{\tabcolsep}{2.5pt}
    \centering
\begin{tabular}{lllllllllllll}
\toprule
 & dataset & Caltech101 & DescribTextures & EuroSAT & FGVCAircraft & Food101 & ImageNet & OxfordFlowers & OxfordPets & SUN397 & StanfordCars & UCF101 \\
num shots & method &  &  &  &  &  &  &  &  &  &  &  \\
\midrule
\multirow[t]{8}{*}{1} & LP$_\text{ICML'21}$ & {\color{red} \ding{55}} & {\color{red} \ding{55}} & {\color{red} \ding{55}} & {\color{red} \ding{55}} & {\color{red} \ding{55}} & {\color{red} \ding{55}} & {\color{red} \ding{55}} & {\color{red} \ding{55}} & {\color{red} \ding{55}} & {\color{red} \ding{55}} & {\color{red} \ding{55}} \\
 & TipA$_\text{ECCV'22}$ & $46.30_{\pm 0.21}$ & $3.27_{\pm 0.02}$ & {\color{red} \ding{55}} & {\color{red} \ding{55}} & $19.75_{\pm 0.10}$ & $9.92_{\pm 0.02}$ & $9.57_{\pm 0.10}$ & $20.40_{\pm 0.14}$ & $5.24_{\pm 0.02}$ & $5.15_{\pm 0.07}$ & $8.50_{\pm 0.36}$ \\
 & TipA-f-$_\text{ECCV'22}$ & $46.88_{\pm 0.34}$ & $2.86_{\pm 0.11}$ & $0.00_{\pm 0.00}$ & $0.81_{\pm 0.06}$ & $20.03_{\pm 0.05}$ & $9.94_{\pm 0.00}$ & $10.35_{\pm 0.09}$ & $21.20_{\pm 0.22}$ & $5.27_{\pm 0.02}$ & $5.29_{\pm 0.06}$ & $9.14_{\pm 0.38}$ \\
 & CrossModal$_\text{CVPR'23}$ & $56.08_{\pm 0.54}$ & $7.35_{\pm 0.38}$ & {\color{red} \ding{55}} & $1.59_{\pm 0.12}$ & $25.26_{\pm 0.24}$ & $7.91_{\pm 0.09}$ & $17.35_{\pm 0.38}$ & $31.70_{\pm 0.68}$ & $7.71_{\pm 0.00}$ & $8.44_{\pm 0.13}$ & $15.17_{\pm 0.39}$ \\
 & TaskRes$_\text{CVPR'23}$ & $52.71_{\pm 0.56}$ & $5.02_{\pm 0.39}$ & {\color{red} \ding{55}} & $1.37_{\pm 0.13}$ & $25.10_{\pm 0.33}$ & $13.24_{\pm 0.09}$ & $14.28_{\pm 0.31}$ & $31.12_{\pm 0.55}$ & $6.97_{\pm 0.02}$ & $7.46_{\pm 0.14}$ & $12.35_{\pm 0.51}$ \\
 & LP++$_\text{CVPR'24}$ & $5.21_{\pm 0.39}$ & $0.00_{\pm 0.00}$ & {\color{red} \ding{55}} & $0.20_{\pm 0.13}$ & $1.94_{\pm 0.43}$ & $0.22_{\pm 0.00}$ & $2.50_{\pm 0.57}$ & $4.92_{\pm 1.77}$ & $0.07_{\pm 0.02}$ & $0.71_{\pm 0.22}$ & $0.04_{\pm 0.02}$ \\
\bottomrule
\end{tabular}
    }
    \caption{Selective classification at 95\% confidence, using ResNet-50 backbone. We show the test set coverage provided the accuracy on the selected test set is greater than or equal to the requested confidence. Otherwise, the method is not reliable at such confidence level and we report {\color{red} \ding{55}}.}
    \label{tab:full_SC_rn50_95}
\end{table*}

\begin{table*}[ht]
{\tiny
\setlength{\tabcolsep}{2.5pt}
    \centering
\begin{tabular}{lllllllllllll}
\toprule
 & dataset & Caltech101 & DescribTextures & EuroSAT & FGVCAircraft & Food101 & ImageNet & OxfordFlowers & OxfordPets & SUN397 & StanfordCars & UCF101 \\
num shots & method &  &  &  &  &  &  &  &  &  &  &  \\
\midrule
\multirow[t]{8}{*}{1} & LP$_\text{ICML'21}$ & {\color{red} \ding{55}} & {\color{red} \ding{55}} & {\color{red} \ding{55}} & {\color{red} \ding{55}} & {\color{red} \ding{55}} & {\color{red} \ding{55}} & {\color{red} \ding{55}} & {\color{red} \ding{55}} & {\color{red} \ding{55}} & {\color{red} \ding{55}} & {\color{red} \ding{55}} \\
 & TipA$_\text{ECCV'22}$ & $55.10_{\pm 0.11}$ & $5.34_{\pm 0.12}$ & {\color{red} \ding{55}} & $1.20_{\pm 0.05}$ & $28.25_{\pm 0.03}$ & $15.31_{\pm 0.07}$ & $15.73_{\pm 0.03}$ & $31.73_{\pm 0.10}$ & $9.34_{\pm 0.03}$ & $8.21_{\pm 0.12}$ & $13.69_{\pm 0.37}$ \\
 & TipA-f-$_\text{ECCV'22}$ & $55.70_{\pm 0.15}$ & $4.71_{\pm 0.08}$ & {\color{red} \ding{55}} & $1.35_{\pm 0.06}$ & $28.54_{\pm 0.05}$ & $15.31_{\pm 0.06}$ & $16.94_{\pm 0.19}$ & $32.38_{\pm 0.11}$ & $9.39_{\pm 0.04}$ & $8.46_{\pm 0.12}$ & $14.57_{\pm 0.43}$ \\
 & CrossModal$_\text{CVPR'23}$ & $65.56_{\pm 0.14}$ & $12.29_{\pm 0.67}$ & {\color{red} \ding{55}} & $2.32_{\pm 0.14}$ & $35.04_{\pm 0.26}$ & $13.79_{\pm 0.09}$ & $27.70_{\pm 0.13}$ & $44.14_{\pm 1.02}$ & $14.11_{\pm 0.10}$ & $13.33_{\pm 0.04}$ & $23.87_{\pm 0.29}$ \\
 & TaskRes$_\text{CVPR'23}$ & $62.79_{\pm 0.12}$ & $9.18_{\pm 0.46}$ & {\color{red} \ding{55}} & $1.96_{\pm 0.08}$ & $35.19_{\pm 0.19}$ & $20.36_{\pm 0.08}$ & $23.47_{\pm 0.24}$ & $43.32_{\pm 0.50}$ & $12.80_{\pm 0.12}$ & $11.77_{\pm 0.07}$ & $20.93_{\pm 0.56}$ \\
 & LP++$_\text{CVPR'24}$ & $15.77_{\pm 0.51}$ & $0.08_{\pm 0.08}$ & {\color{red} \ding{55}} & $0.53_{\pm 0.28}$ & $5.73_{\pm 1.08}$ & $0.69_{\pm 0.02}$ & $6.86_{\pm 1.23}$ & $11.73_{\pm 3.12}$ & $0.24_{\pm 0.05}$ & $1.80_{\pm 0.42}$ & $0.33_{\pm 0.11}$ \\
\bottomrule
\end{tabular}
    }
    \caption{Selective classification at 90\% confidence, using ResNet-50 backbone. We show the test set coverage provided the accuracy on the selected test set is greater than or equal to the requested confidence. Otherwise, the method is not reliable at such confidence level and we report {\color{red} \ding{55}}.}
    \label{tab:full_SC_rn50_90}
\end{table*}

\begin{table*}[ht]
{\tiny
\setlength{\tabcolsep}{2.5pt}
    \centering
\begin{tabular}{lllllllllllll}
\toprule
 & dataset & Caltech101 & DescribTextures & EuroSAT & FGVCAircraft & Food101 & ImageNet & OxfordFlowers & OxfordPets & SUN397 & StanfordCars & UCF101 \\
num shots & method &  &  &  &  &  &  &  &  &  &  &  \\
\midrule
\multirow[t]{8}{*}{1} & LP$_\text{ICML'21}$ & {\color{red} \ding{55}} & {\color{red} \ding{55}} & {\color{red} \ding{55}} & {\color{red} \ding{55}} & {\color{red} \ding{55}} & {\color{red} \ding{55}} & {\color{red} \ding{55}} & {\color{red} \ding{55}} & {\color{red} \ding{55}} & {\color{red} \ding{55}} & {\color{red} \ding{55}} \\
 & TipA$_\text{ECCV'22}$ & $59.69_{\pm 0.19}$ & $7.03_{\pm 0.22}$ & {\color{red} \ding{55}} & $1.71_{\pm 0.02}$ & $34.34_{\pm 0.03}$ & $19.54_{\pm 0.09}$ & $20.88_{\pm 0.09}$ & $38.88_{\pm 0.05}$ & $12.65_{\pm 0.03}$ & $10.86_{\pm 0.09}$ & $17.68_{\pm 0.43}$ \\
 & TipA-f-$_\text{ECCV'22}$ & $60.35_{\pm 0.23}$ & $6.13_{\pm 0.15}$ & {\color{red} \ding{55}} & $1.87_{\pm 0.11}$ & $34.56_{\pm 0.03}$ & $19.57_{\pm 0.09}$ & $22.32_{\pm 0.25}$ & $39.79_{\pm 0.13}$ & $12.68_{\pm 0.03}$ & $11.06_{\pm 0.07}$ & $18.77_{\pm 0.49}$ \\
 & CrossModal$_\text{CVPR'23}$ & $71.14_{\pm 0.32}$ & $16.73_{\pm 0.72}$ & {\color{red} \ding{55}} & $2.97_{\pm 0.15}$ & $42.11_{\pm 0.27}$ & $18.66_{\pm 0.12}$ & $34.88_{\pm 0.33}$ & $52.79_{\pm 1.11}$ & $19.77_{\pm 0.20}$ & $17.89_{\pm 0.04}$ & $29.33_{\pm 0.27}$ \\
 & TaskRes$_\text{CVPR'23}$ & $68.82_{\pm 0.24}$ & $13.12_{\pm 0.75}$ & $22.14_{\pm 0.91}$ & $2.55_{\pm 0.11}$ & $42.25_{\pm 0.09}$ & $25.90_{\pm 0.14}$ & $30.34_{\pm 0.39}$ & $51.82_{\pm 0.73}$ & $18.35_{\pm 0.17}$ & $15.86_{\pm 0.03}$ & $26.94_{\pm 0.45}$ \\
 & LP++$_\text{CVPR'24}$ & $27.25_{\pm 1.46}$ & $0.30_{\pm 0.15}$ & {\color{red} \ding{55}} & $0.87_{\pm 0.42}$ & $9.71_{\pm 1.47}$ & $1.35_{\pm 0.05}$ & $10.65_{\pm 1.71}$ & $18.36_{\pm 3.75}$ & $0.49_{\pm 0.11}$ & $3.02_{\pm 0.51}$ & $1.08_{\pm 0.31}$ \\
\bottomrule
\end{tabular}
    }
    \caption{Selective classification at 85\% confidence, using ResNet-50 backbone. We show the test set coverage provided the accuracy on the selected test set is greater than or equal to the requested confidence. Otherwise, the method is not reliable at such confidence level and we report {\color{red} \ding{55}}.}
    \label{tab:full_SC_rn50_85}
\end{table*}

\begin{table*}[ht]
{\tiny
\setlength{\tabcolsep}{2.5pt}
    \centering
\begin{tabular}{lllllllllllll}
\toprule
 & dataset & Caltech101 & DescribTextures & EuroSAT & FGVCAircraft & Food101 & ImageNet & OxfordFlowers & OxfordPets & SUN397 & StanfordCars & UCF101 \\
num shots & method &  &  &  &  &  &  &  &  &  &  &  \\
\midrule
\multirow[t]{8}{*}{1} & LP$_\text{ICML'21}$ & {\color{red} \ding{55}} & {\color{red} \ding{55}} & {\color{red} \ding{55}} & {\color{red} \ding{55}} & {\color{red} \ding{55}} & {\color{red} \ding{55}} & {\color{red} \ding{55}} & {\color{red} \ding{55}} & {\color{red} \ding{55}} & {\color{red} \ding{55}} & {\color{red} \ding{55}} \\
 & TipA$_\text{ECCV'22}$ & $63.23_{\pm 0.31}$ & $8.71_{\pm 0.17}$ & {\color{red} \ding{55}} & $2.15_{\pm 0.07}$ & $39.12_{\pm 0.00}$ & $23.18_{\pm 0.07}$ & $25.61_{\pm 0.11}$ & $44.52_{\pm 0.16}$ & $16.01_{\pm 0.05}$ & $13.07_{\pm 0.13}$ & $21.32_{\pm 0.58}$ \\
 & TipA-f-$_\text{ECCV'22}$ & $63.94_{\pm 0.37}$ & $7.64_{\pm 0.26}$ & {\color{red} \ding{55}} & $2.37_{\pm 0.13}$ & $39.33_{\pm 0.04}$ & $23.21_{\pm 0.06}$ & $26.88_{\pm 0.24}$ & $45.03_{\pm 0.21}$ & $15.98_{\pm 0.07}$ & $13.35_{\pm 0.16}$ & $22.42_{\pm 0.44}$ \\
 & CrossModal$_\text{CVPR'23}$ & $75.15_{\pm 0.17}$ & $20.11_{\pm 0.94}$ & {\color{red} \ding{55}} & $3.94_{\pm 0.10}$ & $47.71_{\pm 0.20}$ & $23.24_{\pm 0.16}$ & $41.85_{\pm 0.20}$ & $59.56_{\pm 1.00}$ & $24.76_{\pm 0.23}$ & $22.47_{\pm 0.15}$ & $34.35_{\pm 0.47}$ \\
 & TaskRes$_\text{CVPR'23}$ & $73.02_{\pm 0.31}$ & $16.53_{\pm 0.73}$ & $29.81_{\pm 0.26}$ & $3.25_{\pm 0.16}$ & $47.95_{\pm 0.07}$ & $30.87_{\pm 0.16}$ & $36.47_{\pm 0.19}$ & $58.80_{\pm 0.65}$ & $23.30_{\pm 0.23}$ & $19.97_{\pm 0.13}$ & $32.08_{\pm 0.41}$ \\
 & LP++$_\text{CVPR'24}$ & $36.34_{\pm 2.12}$ & $0.67_{\pm 0.21}$ & $16.00_{\pm 8.14}$ & $1.12_{\pm 0.52}$ & $13.69_{\pm 1.71}$ & $2.22_{\pm 0.07}$ & $15.39_{\pm 2.46}$ & $24.52_{\pm 3.74}$ & $0.94_{\pm 0.21}$ & $4.19_{\pm 0.60}$ & $1.97_{\pm 0.45}$ \\
\bottomrule
\end{tabular}
    }
    \caption{Selective classification at 80\% confidence, using ResNet-50 backbone. We show the test set coverage provided the accuracy on the selected test set is greater than or equal to the requested confidence. Otherwise, the method is not reliable at such confidence level and we report {\color{red} \ding{55}}.}
    \label{tab:full_SC_rn50_80}
\end{table*}

\begin{table*}[ht]
{\tiny
\setlength{\tabcolsep}{2.5pt}
    \centering

\begin{tabular}{lllllllllllll}
\toprule
 & dataset & Caltech101 & DescribTextures & EuroSAT & FGVCAircraft & Food101 & ImageNet & OxfordFlowers & OxfordPets & SUN397 & StanfordCars & UCF101 \\
num shots & method &  &  &  &  &  &  &  &  &  &  &  \\
\midrule
\multirow[t]{8}{*}{1} & LP$_\text{ICML'21}$ & $76.44_{\pm 0.32}$ & $33.61_{\pm 0.32}$ & $46.12_{\pm 5.68}$ & $17.61_{\pm 0.62}$ & $43.16_{\pm 0.47}$ & $25.54_{\pm 0.21}$ & $65.11_{\pm 0.84}$ & $40.37_{\pm 1.21}$ & $33.13_{\pm 0.78}$ & $28.34_{\pm 0.87}$ & $45.19_{\pm 0.99}$ \\
 & TipA$_\text{ECCV'22}$ & $87.67_{\pm 0.64}$ & $40.52_{\pm 0.62}$ & $49.51_{\pm 2.80}$ & $24.38_{\pm 0.54}$ & $66.54_{\pm 1.11}$ & $49.11_{\pm 0.21}$ & $78.06_{\pm 1.59}$ & $53.77_{\pm 1.33}$ & $52.94_{\pm 0.48}$ & $49.48_{\pm 0.36}$ & $64.68_{\pm 0.44}$ \\
 & TipA-f-$_\text{ECCV'22}$ & $87.67_{\pm 0.64}$ & $40.52_{\pm 0.62}$ & $49.51_{\pm 2.80}$ & $24.38_{\pm 0.54}$ & $66.54_{\pm 1.11}$ & $49.11_{\pm 0.21}$ & $78.06_{\pm 1.59}$ & $53.77_{\pm 1.33}$ & $52.94_{\pm 0.48}$ & $49.48_{\pm 0.36}$ & $64.68_{\pm 0.44}$ \\
 & CrossModal$_\text{CVPR'23}$ & $92.91_{\pm 0.01}$ & $54.29_{\pm 0.75}$ & $58.92_{\pm 0.50}$ & $29.78_{\pm 0.39}$ & $83.50_{\pm 0.22}$ & $63.98_{\pm 0.09}$ & $87.31_{\pm 0.13}$ & $88.44_{\pm 0.36}$ & $67.91_{\pm 0.14}$ & $67.60_{\pm 0.30}$ & $72.59_{\pm 0.49}$ \\
 & TaskRes$_\text{CVPR'23}$ & $93.18_{\pm 0.28}$ & $54.91_{\pm 1.09}$ & $60.97_{\pm 1.29}$ & $30.41_{\pm 0.52}$ & $84.19_{\pm 0.17}$ & $68.97_{\pm 0.08}$ & $85.34_{\pm 0.30}$ & $89.46_{\pm 0.21}$ & $67.67_{\pm 0.09}$ & $68.84_{\pm 0.10}$ & $73.02_{\pm 0.37}$ \\
 & LP++$_\text{CVPR'24}$ & $93.12_{\pm 0.72}$ & $52.29_{\pm 0.52}$ & $66.72_{\pm 1.12}$ & $29.54_{\pm 0.42}$ & $84.46_{\pm 0.93}$ & $61.04_{\pm 0.06}$ & $86.17_{\pm 0.21}$ & $88.70_{\pm 0.81}$ & $67.46_{\pm 0.14}$ & $67.76_{\pm 0.11}$ & $72.68_{\pm 0.29}$ \\
\bottomrule
\end{tabular}
}
    \caption{Accuracy with ViT-16 backbone: full numerical results.}
    \label{tab:full_acc_vit16}
\end{table*}

\begin{table*}[ht]
{\tiny
\setlength{\tabcolsep}{2.5pt}
    \centering
    \begin{tabular}{lllllllllllll}
\toprule
 & dataset & Caltech101 & DescribTextures & EuroSAT & FGVCAircraft & Food101 & ImageNet & OxfordFlowers & OxfordPets & SUN397 & StanfordCars & UCF101 \\
num shots & method &  &  &  &  &  &  &  &  &  &  &  \\
\midrule
\multirow[t]{8}{*}{1} & LP$_\text{ICML'21}$ & $15.56_{\pm 0.49}$ & $50.41_{\pm 1.12}$ & $40.46_{\pm 4.02}$ & $57.50_{\pm 0.63}$ & $39.21_{\pm 0.21}$ & $42.02_{\pm 0.08}$ & $19.04_{\pm 0.35}$ & $41.00_{\pm 1.19}$ & $45.07_{\pm 0.64}$ & $46.45_{\pm 1.62}$ & $39.67_{\pm 1.53}$ \\
 & TipA$_\text{ECCV'22}$ & $9.49_{\pm 0.59}$ & $50.98_{\pm 0.58}$ & $44.52_{\pm 3.71}$ & $47.19_{\pm 0.59}$ & $25.20_{\pm 0.97}$ & $38.49_{\pm 0.17}$ & $18.34_{\pm 1.51}$ & $41.30_{\pm 1.07}$ & $35.21_{\pm 0.56}$ & $37.72_{\pm 0.43}$ & $23.95_{\pm 0.37}$ \\
 & TipA-f-$_\text{ECCV'22}$ & $9.49_{\pm 0.59}$ & $50.98_{\pm 0.58}$ & $44.52_{\pm 3.71}$ & $47.19_{\pm 0.59}$ & $25.20_{\pm 0.97}$ & $38.49_{\pm 0.17}$ & $18.34_{\pm 1.51}$ & $41.30_{\pm 1.07}$ & $35.21_{\pm 0.56}$ & $37.72_{\pm 0.43}$ & $23.95_{\pm 0.37}$ \\
 & CrossModal$_\text{CVPR'23}$ & $1.81_{\pm 0.29}$ & $3.27_{\pm 0.97}$ & $13.00_{\pm 1.61}$ & $10.25_{\pm 0.11}$ & $1.52_{\pm 0.12}$ & $7.58_{\pm 0.14}$ & $10.82_{\pm 0.32}$ & $2.41_{\pm 0.50}$ & $2.93_{\pm 0.14}$ & $1.56_{\pm 0.20}$ & $2.16_{\pm 0.60}$ \\
 & TaskRes$_\text{CVPR'23}$ & $2.81_{\pm 0.15}$ & $3.10_{\pm 0.88}$ & $7.44_{\pm 2.52}$ & $5.12_{\pm 0.32}$ & $2.27_{\pm 0.14}$ & $1.65_{\pm 0.12}$ & $12.55_{\pm 0.51}$ & $3.74_{\pm 0.32}$ & $3.90_{\pm 0.09}$ & $4.55_{\pm 0.13}$ & $2.07_{\pm 0.37}$ \\
 & LP++$_\text{CVPR'24}$ & $8.31_{\pm 1.29}$ & $8.25_{\pm 3.52}$ & $17.53_{\pm 6.24}$ & $4.00_{\pm 1.20}$ & $6.78_{\pm 1.00}$ & $25.95_{\pm 0.23}$ & $14.52_{\pm 0.37}$ & $6.99_{\pm 1.52}$ & $13.53_{\pm 2.03}$ & $11.12_{\pm 0.29}$ & $6.73_{\pm 1.54}$ \\
\bottomrule
\end{tabular}
}
    \caption{ECE with ViT-16 backbone: full numerical results.}
    \label{tab:full_ece_vit16}
\end{table*}

\begin{table*}[ht]
{\tiny
\setlength{\tabcolsep}{2.5pt}
    \centering
\begin{tabular}{lllllllllllll}
\toprule
 & dataset & Caltech101 & DescribTextures & EuroSAT & FGVCAircraft & Food101 & ImageNet & OxfordFlowers & OxfordPets & SUN397 & StanfordCars & UCF101 \\
num shots & method &  &  &  &  &  &  &  &  &  &  &  \\
\midrule
\multirow[t]{8}{*}{1} & LP$_\text{ICML'21}$ & $15.55_{\pm 0.49}$ & $50.40_{\pm 1.12}$ & $40.45_{\pm 4.03}$ & $57.49_{\pm 0.62}$ & $39.21_{\pm 0.21}$ & $42.02_{\pm 0.08}$ & $18.99_{\pm 0.31}$ & $41.00_{\pm 1.18}$ & $45.07_{\pm 0.64}$ & $46.45_{\pm 1.62}$ & $39.66_{\pm 1.53}$ \\
 & TipA$_\text{ECCV'22}$ & $9.32_{\pm 0.50}$ & $50.92_{\pm 0.56}$ & $44.51_{\pm 3.71}$ & $47.19_{\pm 0.59}$ & $25.20_{\pm 0.97}$ & $38.47_{\pm 0.18}$ & $18.32_{\pm 1.51}$ & $41.28_{\pm 1.07}$ & $35.21_{\pm 0.56}$ & $37.71_{\pm 0.43}$ & $23.93_{\pm 0.37}$ \\
 & TipA-f-$_\text{ECCV'22}$ & $9.32_{\pm 0.50}$ & $50.92_{\pm 0.56}$ & $44.51_{\pm 3.71}$ & $47.19_{\pm 0.59}$ & $25.20_{\pm 0.97}$ & $38.47_{\pm 0.18}$ & $18.32_{\pm 1.51}$ & $41.28_{\pm 1.07}$ & $35.21_{\pm 0.56}$ & $37.71_{\pm 0.43}$ & $23.93_{\pm 0.37}$ \\
 & CrossModal$_\text{CVPR'23}$ & $1.36_{\pm 0.32}$ & $3.56_{\pm 0.86}$ & $13.00_{\pm 1.61}$ & $10.19_{\pm 0.13}$ & $1.52_{\pm 0.13}$ & $7.62_{\pm 0.13}$ & $10.81_{\pm 0.30}$ & $2.45_{\pm 0.54}$ & $2.98_{\pm 0.14}$ & $1.64_{\pm 0.05}$ & $2.05_{\pm 0.61}$ \\
 & TaskRes$_\text{CVPR'23}$ & $2.75_{\pm 0.16}$ & $3.24_{\pm 0.81}$ & $7.46_{\pm 2.54}$ & $4.97_{\pm 0.39}$ & $2.27_{\pm 0.13}$ & $1.63_{\pm 0.08}$ & $12.55_{\pm 0.51}$ & $3.66_{\pm 0.40}$ & $3.96_{\pm 0.06}$ & $4.46_{\pm 0.15}$ & $1.98_{\pm 0.36}$ \\
 & LP++$_\text{CVPR'24}$ & $8.29_{\pm 1.28}$ & $8.41_{\pm 3.38}$ & $17.49_{\pm 6.28}$ & $4.33_{\pm 0.94}$ & $6.78_{\pm 1.00}$ & $25.96_{\pm 0.23}$ & $14.51_{\pm 0.36}$ & $6.98_{\pm 1.50}$ & $13.53_{\pm 2.03}$ & $11.11_{\pm 0.30}$ & $6.73_{\pm 1.53}$ \\
\bottomrule
\end{tabular}
}
    \caption{AECE with ViT-16 backbone: full numerical results.}
    \label{tab:full_aece_vit16}
\end{table*}

\begin{table*}[ht]
{\tiny
\setlength{\tabcolsep}{2.5pt}
    \centering
\begin{tabular}{lllllllllllll}
\toprule
 & dataset & Caltech101 & DescribTextures & EuroSAT & FGVCAircraft & Food101 & ImageNet & OxfordFlowers & OxfordPets & SUN397 & StanfordCars & UCF101 \\
num shots & method &  &  &  &  &  &  &  &  &  &  &  \\
\midrule
\multirow[t]{8}{*}{1} & LP$_\text{ICML'21}$ & {\color{red} \ding{55}} & {\color{red} \ding{55}} & {\color{red} \ding{55}} & {\color{red} \ding{55}} & {\color{red} \ding{55}} & {\color{red} \ding{55}} & {\color{red} \ding{55}} & {\color{red} \ding{55}} & {\color{red} \ding{55}} & {\color{red} \ding{55}} & {\color{red} \ding{55}} \\
 & TipA$_\text{ECCV'22}$ & {\color{red} \ding{55}} & {\color{red} \ding{55}} & {\color{red} \ding{55}} & {\color{red} \ding{55}} & {\color{red} \ding{55}} & {\color{red} \ding{55}} & {\color{red} \ding{55}} & {\color{red} \ding{55}} & {\color{red} \ding{55}} & {\color{red} \ding{55}} & {\color{red} \ding{55}} \\
 & TipA-f-$_\text{ECCV'22}$ & {\color{red} \ding{55}} & {\color{red} \ding{55}} & {\color{red} \ding{55}} & {\color{red} \ding{55}} & {\color{red} \ding{55}} & {\color{red} \ding{55}} & {\color{red} \ding{55}} & {\color{red} \ding{55}} & {\color{red} \ding{55}} & {\color{red} \ding{55}} & {\color{red} \ding{55}} \\
 & CrossModal$_\text{CVPR'23}$ & $42.38_{\pm 1.28}$ & $3.49_{\pm 0.29}$ & {\color{red} \ding{55}} & $1.35_{\pm 0.27}$ & $22.09_{\pm 0.37}$ & {\color{red} \ding{55}} & $8.81_{\pm 0.55}$ & $21.25_{\pm 0.42}$ & $3.47_{\pm 0.07}$ & $6.86_{\pm 0.14}$ & $10.96_{\pm 0.92}$ \\
 & TaskRes$_\text{CVPR'23}$ & $38.97_{\pm 1.26}$ & $2.32_{\pm 0.07}$ & {\color{red} \ding{55}} & $1.11_{\pm 0.27}$ & $21.58_{\pm 0.35}$ & {\color{red} \ding{55}} & $6.71_{\pm 0.55}$ & $20.11_{\pm 0.78}$ & $3.09_{\pm 0.10}$ & $6.25_{\pm 0.18}$ & $9.12_{\pm 0.81}$ \\
 & LP++$_\text{CVPR'24}$ & $5.53_{\pm 0.45}$ & $0.02_{\pm 0.02}$ & $0.48_{\pm 0.48}$ & $0.09_{\pm 0.00}$ & $5.84_{\pm 1.08}$ & $0.01_{\pm 0.00}$ & $1.30_{\pm 0.20}$ & $4.92_{\pm 1.01}$ & $0.11_{\pm 0.05}$ & $0.82_{\pm 0.16}$ & $0.86_{\pm 0.23}$ \\
\end{tabular}
}
     \caption{Selective classification at 99\% confidence, using ViT-16 backbone. We show the test set coverage provided the accuracy on the selected test set is greater than or equal to the requested confidence. Otherwise, the method is not reliable at such confidence level and we report {\color{red} \ding{55}}.}
    \label{tab:full_SC_vit16_99}
\end{table*}

\begin{table*}[ht]
{\tiny
\setlength{\tabcolsep}{2.5pt}
    \centering
\begin{tabular}{lllllllllllll}
\toprule
 & dataset & Caltech101 & DescribTextures & EuroSAT & FGVCAircraft & Food101 & ImageNet & OxfordFlowers & OxfordPets & SUN397 & StanfordCars & UCF101 \\
num shots & method &  &  &  &  &  &  &  &  &  &  &  \\
\midrule
\multirow[t]{8}{*}{1} & LP$_\text{ICML'21}$ & {\color{red} \ding{55}} & {\color{red} \ding{55}} & {\color{red} \ding{55}} & {\color{red} \ding{55}} & {\color{red} \ding{55}} & {\color{red} \ding{55}} & {\color{red} \ding{55}} & {\color{red} \ding{55}} & {\color{red} \ding{55}} & {\color{red} \ding{55}} & {\color{red} \ding{55}} \\
 & TipA$_\text{ECCV'22}$ & {\color{red} \ding{55}} & {\color{red} \ding{55}} & {\color{red} \ding{55}} & {\color{red} \ding{55}} & {\color{red} \ding{55}} & {\color{red} \ding{55}} & {\color{red} \ding{55}} & {\color{red} \ding{55}} & {\color{red} \ding{55}} & {\color{red} \ding{55}} & {\color{red} \ding{55}} \\
 & TipA-f-$_\text{ECCV'22}$ & {\color{red} \ding{55}} & {\color{red} \ding{55}} & {\color{red} \ding{55}} & {\color{red} \ding{55}} & {\color{red} \ding{55}} & {\color{red} \ding{55}} & {\color{red} \ding{55}} & {\color{red} \ding{55}} & {\color{red} \ding{55}} & {\color{red} \ding{55}} & {\color{red} \ding{55}} \\
 & CrossModal$_\text{CVPR'23}$ & $66.99_{\pm 0.53}$ & $11.68_{\pm 0.50}$ & {\color{red} \ding{55}} & $3.80_{\pm 0.18}$ & $43.78_{\pm 0.28}$ & $8.10_{\pm 0.03}$ & $27.31_{\pm 0.22}$ & $47.20_{\pm 0.81}$ & $13.71_{\pm 0.12}$ & $16.72_{\pm 0.04}$ & $30.03_{\pm 0.57}$ \\
 & TaskRes$_\text{CVPR'23}$ & $63.23_{\pm 1.10}$ & $8.96_{\pm 0.29}$ & {\color{red} \ding{55}} & $3.28_{\pm 0.19}$ & $43.09_{\pm 0.26}$ & $22.54_{\pm 0.09}$ & $22.06_{\pm 0.55}$ & $46.72_{\pm 0.25}$ & $12.45_{\pm 0.07}$ & $15.17_{\pm 0.17}$ & $26.52_{\pm 0.32}$ \\
 & LP++$_\text{CVPR'24}$ & $28.29_{\pm 3.32}$ & $1.26_{\pm 0.76}$ & $3.31_{\pm 3.31}$ & $1.08_{\pm 0.26}$ & $22.04_{\pm 1.89}$ & $0.22_{\pm 0.00}$ & $11.64_{\pm 0.91}$ & $21.97_{\pm 2.32}$ & $1.63_{\pm 0.55}$ & $4.68_{\pm 0.31}$ & $9.58_{\pm 1.47}$ \\
\bottomrule
\end{tabular}
}
    \caption{Selective classification at 95\% confidence, using ViT-16 backbone. We show the test set coverage provided the accuracy on the selected test set is greater than or equal to the requested confidence. Otherwise, the method is not reliable at such confidence level and we report {\color{red} \ding{55}}.}
    \label{tab:full_SC_vit16_95}
\end{table*}

\begin{table*}[ht]
{\tiny
\setlength{\tabcolsep}{2.5pt}
    \centering
\begin{tabular}{lllllllllllll}
\toprule
 & dataset & Caltech101 & DescribTextures & EuroSAT & FGVCAircraft & Food101 & ImageNet & OxfordFlowers & OxfordPets & SUN397 & StanfordCars & UCF101 \\
num shots & method &  &  &  &  &  &  &  &  &  &  &  \\
\midrule
\multirow[t]{8}{*}{1} & LP$_\text{ICML'21}$ & {\color{red} \ding{55}} & {\color{red} \ding{55}} & {\color{red} \ding{55}} & {\color{red} \ding{55}} & {\color{red} \ding{55}} & {\color{red} \ding{55}} & {\color{red} \ding{55}} & {\color{red} \ding{55}} & {\color{red} \ding{55}} & {\color{red} \ding{55}} & {\color{red} \ding{55}} \\
 & TipA$_\text{ECCV'22}$ & $88.82_{\pm 0.55}$ & {\color{red} \ding{55}} & {\color{red} \ding{55}} & {\color{red} \ding{55}} & {\color{red} \ding{55}} & {\color{red} \ding{55}} & {\color{red} \ding{55}} & {\color{red} \ding{55}} & {\color{red} \ding{55}} & {\color{red} \ding{55}} & {\color{red} \ding{55}} \\
 & TipA-f-$_\text{ECCV'22}$ & $88.82_{\pm 0.55}$ & {\color{red} \ding{55}} & {\color{red} \ding{55}} & {\color{red} \ding{55}} & {\color{red} \ding{55}} & {\color{red} \ding{55}} & {\color{red} \ding{55}} & {\color{red} \ding{55}} & {\color{red} \ding{55}} & {\color{red} \ding{55}} & {\color{red} \ding{55}} \\
 & CrossModal$_\text{CVPR'23}$ & $76.06_{\pm 0.88}$ & $17.69_{\pm 1.05}$ & {\color{red} \ding{55}} & $5.38_{\pm 0.37}$ & $54.67_{\pm 0.12}$ & $15.15_{\pm 0.07}$ & $41.28_{\pm 1.06}$ & $60.05_{\pm 1.25}$ & $22.56_{\pm 0.12}$ & $24.20_{\pm 0.20}$ & $40.48_{\pm 0.71}$ \\
 & TaskRes$_\text{CVPR'23}$ & $73.01_{\pm 0.63}$ & $14.38_{\pm 0.70}$ & $23.44_{\pm 1.94}$ & $4.58_{\pm 0.26}$ & $54.42_{\pm 0.01}$ & $31.72_{\pm 0.15}$ & $34.61_{\pm 0.27}$ & $59.00_{\pm 0.93}$ & $20.90_{\pm 0.17}$ & $21.70_{\pm 0.15}$ & $37.53_{\pm 0.21}$ \\
 & LP++$_\text{CVPR'24}$ & $42.72_{\pm 3.38}$ & $3.23_{\pm 1.62}$ & $6.14_{\pm 6.10}$ & $1.91_{\pm 0.22}$ & $33.04_{\pm 2.17}$ & $0.69_{\pm 0.02}$ & $20.07_{\pm 0.95}$ & $35.50_{\pm 1.80}$ & $4.19_{\pm 1.22}$ & $8.29_{\pm 0.32}$ & $18.35_{\pm 2.79}$ \\
\bottomrule
\end{tabular}
}
    \caption{Selective classification at 90\% confidence, using ViT-16 backbone. We show the test set coverage provided the accuracy on the selected test set is greater than or equal to the requested confidence. Otherwise, the method is not reliable at such confidence level and we report {\color{red} \ding{55}}.}
    \label{tab:full_SC_vit16_90}
\end{table*}

\begin{table*}[ht]
{\tiny
\setlength{\tabcolsep}{2.5pt}
    \centering
\begin{tabular}{lllllllllllll}
\toprule
 & dataset & Caltech101 & DescribTextures & EuroSAT & FGVCAircraft & Food101 & ImageNet & OxfordFlowers & OxfordPets & SUN397 & StanfordCars & UCF101 \\
num shots & method &  &  &  &  &  &  &  &  &  &  &  \\
\midrule
\multirow[t]{8}{*}{1} & LP$_\text{ICML'21}$ & $81.18_{\pm 1.27}$ & {\color{red} \ding{55}} & {\color{red} \ding{55}} & {\color{red} \ding{55}} & {\color{red} \ding{55}} & {\color{red} \ding{55}} & {\color{red} \ding{55}} & {\color{red} \ding{55}} & {\color{red} \ding{55}} & {\color{red} \ding{55}} & {\color{red} \ding{55}} \\
 & TipA$_\text{ECCV'22}$ & $90.57_{\pm 0.62}$ & {\color{red} \ding{55}} & {\color{red} \ding{55}} & {\color{red} \ding{55}} & {\color{red} \ding{55}} & {\color{red} \ding{55}} & {\color{red} \ding{55}} & {\color{red} \ding{55}} & {\color{red} \ding{55}} & {\color{red} \ding{55}} & {\color{red} \ding{55}} \\
 & TipA-f-$_\text{ECCV'22}$ & $90.57_{\pm 0.62}$ & {\color{red} \ding{55}} & {\color{red} \ding{55}} & {\color{red} \ding{55}} & {\color{red} \ding{55}} & {\color{red} \ding{55}} & {\color{red} \ding{55}} & {\color{red} \ding{55}} & {\color{red} \ding{55}} & {\color{red} \ding{55}} & {\color{red} \ding{55}} \\
 & CrossModal$_\text{CVPR'23}$ & $81.76_{\pm 0.88}$ & $22.83_{\pm 1.54}$ & {\color{red} \ding{55}} & $6.94_{\pm 0.50}$ & $61.46_{\pm 0.22}$ & $21.12_{\pm 0.11}$ & $50.17_{\pm 1.13}$ & $67.85_{\pm 1.43}$ & $29.41_{\pm 0.11}$ & $30.29_{\pm 0.52}$ & $47.32_{\pm 0.82}$ \\
 & TaskRes$_\text{CVPR'23}$ & $78.58_{\pm 0.43}$ & $18.99_{\pm 1.10}$ & $30.65_{\pm 2.69}$ & $5.60_{\pm 0.30}$ & $61.34_{\pm 0.09}$ & $38.54_{\pm 0.16}$ & $43.17_{\pm 0.91}$ & $66.93_{\pm 0.91}$ & $27.56_{\pm 0.07}$ & $27.29_{\pm 0.21}$ & $44.80_{\pm 0.63}$ \\
 & LP++$_\text{CVPR'24}$ & $52.17_{\pm 3.66}$ & $5.83_{\pm 2.42}$ & $8.42_{\pm 8.21}$ & $2.71_{\pm 0.32}$ & $40.42_{\pm 2.29}$ & $1.35_{\pm 0.05}$ & $27.50_{\pm 0.90}$ & $44.56_{\pm 1.62}$ & $7.04_{\pm 1.68}$ & $11.11_{\pm 0.43}$ & $25.12_{\pm 3.14}$ \\
\bottomrule
\end{tabular}
}
   \caption{Selective classification at 85\% confidence, using ViT-16 backbone. We show the test set coverage provided the accuracy on the selected test set is greater than or equal to the requested confidence. Otherwise, the method is not reliable at such confidence level and we report {\color{red} \ding{55}}.}
    \label{tab:full_SC_vit16_85}
\end{table*}

\begin{table*}[ht]
{\tiny
\setlength{\tabcolsep}{2.5pt}
    \centering
\begin{tabular}{lllllllllllll}
\toprule
 & dataset & Caltech101 & DescribTextures & EuroSAT & FGVCAircraft & Food101 & ImageNet & OxfordFlowers & OxfordPets & SUN397 & StanfordCars & UCF101 \\
num shots & method &  &  &  &  &  &  &  &  &  &  &  \\
\midrule
\multirow[t]{8}{*}{1} & LP$_\text{ICML'21}$ & $83.80_{\pm 1.31}$ & {\color{red} \ding{55}} & {\color{red} \ding{55}} & {\color{red} \ding{55}} & {\color{red} \ding{55}} & {\color{red} \ding{55}} & $68.75_{\pm 1.27}$ & {\color{red} \ding{55}} & {\color{red} \ding{55}} & {\color{red} \ding{55}} & {\color{red} \ding{55}} \\
 & TipA$_\text{ECCV'22}$ & $91.85_{\pm 0.65}$ & {\color{red} \ding{55}} & {\color{red} \ding{55}} & {\color{red} \ding{55}} & {\color{red} \ding{55}} & {\color{red} \ding{55}} & $90.19_{\pm 0.32}$ & {\color{red} \ding{55}} & {\color{red} \ding{55}} & {\color{red} \ding{55}} & {\color{red} \ding{55}} \\
 & TipA-f-$_\text{ECCV'22}$ & $91.85_{\pm 0.65}$ & {\color{red} \ding{55}} & {\color{red} \ding{55}} & {\color{red} \ding{55}} & {\color{red} \ding{55}} & {\color{red} \ding{55}} & $90.19_{\pm 0.32}$ & {\color{red} \ding{55}} & {\color{red} \ding{55}} & {\color{red} \ding{55}} & {\color{red} \ding{55}} \\
 & CrossModal$_\text{CVPR'23}$ & $85.41_{\pm 0.78}$ & $27.50_{\pm 1.96}$ & $43.06_{\pm 2.65}$ & $8.67_{\pm 0.60}$ & $66.54_{\pm 0.24}$ & $26.36_{\pm 0.08}$ & $57.42_{\pm 0.61}$ & $73.44_{\pm 1.22}$ & $35.61_{\pm 0.09}$ & $36.09_{\pm 0.64}$ & $52.84_{\pm 0.72}$ \\
 & TaskRes$_\text{CVPR'23}$ & $82.58_{\pm 0.59}$ & $23.07_{\pm 1.59}$ & $36.99_{\pm 3.01}$ & $6.76_{\pm 0.46}$ & $66.47_{\pm 0.12}$ & $44.16_{\pm 0.12}$ & $50.52_{\pm 0.99}$ & $73.11_{\pm 1.03}$ & $33.58_{\pm 0.11}$ & $32.60_{\pm 0.13}$ & $50.12_{\pm 0.77}$ \\
 & LP++$_\text{CVPR'24}$ & $58.66_{\pm 3.68}$ & $7.88_{\pm 2.85}$ & $10.68_{\pm 10.05}$ & $3.40_{\pm 0.30}$ & $46.25_{\pm 2.26}$ & $2.22_{\pm 0.07}$ & $33.96_{\pm 0.86}$ & $51.20_{\pm 1.10}$ & $10.18_{\pm 2.07}$ & $13.98_{\pm 0.46}$ & $30.67_{\pm 2.93}$ \\
\bottomrule
\end{tabular}
}
    \caption{Selective classification at 80\% confidence, using ViT-16 backbone. We show the test set coverage provided the accuracy on the selected test set is greater than or equal to the requested confidence. Otherwise, the method is not reliable at such confidence level and we report {\color{red} \ding{55}}.}
    \label{tab:full_SC_vit16_80}
\end{table*}

\end{document}